\PassOptionsToPackage{hyphens}{url}
\PassOptionsToPackage{table}{xcolor}
\documentclass[]{academic_template}

\usepackage[toc,page,header]{appendix}
\usepackage{amssymb}
\usepackage{amsmath}
\usepackage{makecell}
\usepackage{graphicx}
\usepackage{listings}
\usepackage[hyphens]{url}
\usepackage{booktabs}
\usepackage{colortbl}
\usepackage[table]{xcolor}
\usepackage{caption}
\usepackage{subcaption}

\graphicspath{{figures/paper/}}
\tcbuselibrary{listings}
\tcbset{
  aibox/.style={
    width=0.94\linewidth,
    top=10pt,
    bottom=7pt,
    left=9pt,
    right=9pt,
    colback=white,
    colframe=black,
    colbacktitle=black,
    coltitle=white,
    boxrule=0.5pt,
    enhanced,
    center,
    breakable,
    attach boxed title to top left={yshift=-0.1in,xshift=0.15in},
    boxed title style={boxrule=0pt,colframe=white},
    fonttitle=\bfseries\small,
  }
}
\newtcblisting{AIbox}[1]{
  aibox,
  title={#1},
  listing only,
  listing engine=listings,
  listing options={
    basicstyle=\footnotesize\ttfamily,
    breaklines=true,
    breakatwhitespace=false,
    columns=fullflexible,
    keepspaces=true,
    showstringspaces=false,
    tabsize=2
  }
}

\setlogoheight{14mm}
\setlogospacing{5mm}
\setsjtublue
\settitlerulethickness{3pt}
\abstractboxon
\setabstractframecolor{gray}
\setabstractbgcolor{gray!10}
\setlogotolineshift{5mm}
\settoprulethickness{2.5pt}
\setbottomrulethickness{1.5pt}

\title{FIRM-Video: Check Before You Score for Reliable Text-to-Video Reward Modeling}

\author[1,2]{Peiyuan Zhang}
\author[1]{Xiangyu Zhao}
\author[3,2]{Hongbo liu}
\author[1,2]{Xiaoxing Hu}
\author[1,2]{Mingxin Liu}
\author[1]{Shuran Ma}
\author[2]{Yunhang Shen}
\author[2]{Jian Hu}
\author[2]{Haihan Gao}
\author[2, \dagger]{Haoyu Cao}
\author[1, \dagger]{Xue Yang}

\affiliation[1]{Shanghai Jiao Tong University}
\affiliation[2]{Tencent Youtu Lab}
\affiliation[3]{Tongji University}

\contribution[\dagger]{Corresponding Author}

\abstract{
Reliable reward models are essential for text-to-video evaluation and alignment. However, the trade-off between evaluation accuracy and inference efficiency places high demands on the quality of training supervision. Existing approaches often rely on holistic judges with fixed rubrics or open-ended reasoning, leading to incomplete inspection, unfaithful justification, and entangled attribution. We introduce \textbf{FIRM-Video}, a unified checklist-driven data construction framework based on a \textit{check-before-score} principle: construct dimension-specific checklists, verify each criterion against temporal visual evidence, and aggregate only verified decisions. For Instruction Following, FIRM-Video decomposes prompts into weighted atomic requirements; for World Coherence, it constructs prompt-calibrated, target-specific checks grounded in visible entities and actions; and for Perceptual Quality, it applies a generic taxonomy of visual defects. The verified criteria and scores are further transformed into natural-language analyses for end-to-end reward modeling. Subsequently, we construct FIRM-Video-90K with 88,044 dimension-specific instances from 29,348 videos, and introduce FIRM-Video-Bench with 750 point-wise human annotations across 250 videos. The Qwen3-VL-based FIRM-Video-8B achieves the best overall MAE on FIRM-Video-Bench while consistently delivering the highest VBench Total, Quality, and Semantic Scores in Best-of-8 sampling across three video generators.
}

\date{\today}
\checkdata[Code]{\url{https://github.com/VisionXLab/FIRM-Reward}}
\checkdata[Hugging Face]{\url{https://huggingface.co/collections/VisionXLab/firm-reward}}

\begin{document}
\maketitle

\section{Introduction}

Recent advances in text-to-video (T2V) generation have substantially improved visual fidelity, motion dynamics, and semantic controllability~\cite{brooks2024video,kong2024hunyuanvideo,wan2025wan}. However, reliably evaluating generated videos remains challenging. A satisfactory video should (i) faithfully realize the user prompt, (ii) maintain physical plausibility in entities, motions, and interactions, and (iii) provide a high-quality viewing experience. We refer to these complementary dimensions as \textbf{Instruction Following (IF)}, \textbf{World Coherence (WC)}, and \textbf{Perceptual Quality (PQ)}, respectively. Reward models that assess these dimensions are central to benchmarking, data filtering, best-of-N sampling, and reinforcement-learning-based alignment~\cite{liu2026improving}. 

\begin{figure}[t]
    \centering
    \includegraphics[width=\columnwidth]{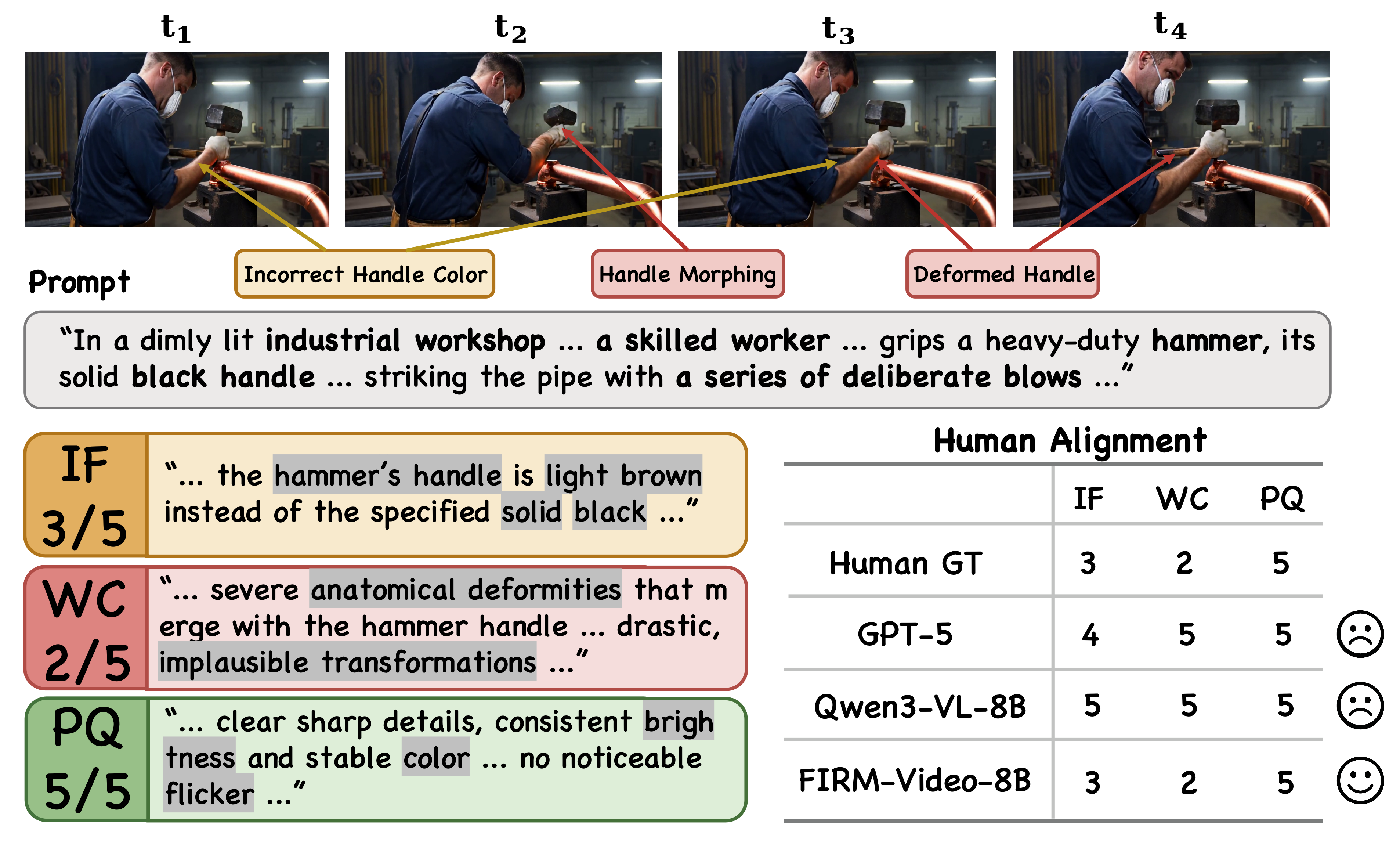}
    \caption{
        Qualitative evaluation case of \textbf{FIRM-Video-8B}, illustrating its improved alignment with human judgments.
    }
    \label{fig:teaser}
\end{figure}

Reliable evaluation is essential for developing and assessing video generation models, yet achieving both reliability and efficiency remains difficult. Complex multi-stage reasoning pipelines can improve evaluation quality, but their high computational cost limits scalability. End-to-end reward models are far more efficient, but their reliability depends heavily on the quality of their training supervision. The key challenge is therefore to construct reliable supervision for end-to-end reward modeling.
Existing approaches typically rely on holistic judges guided by fixed rubrics or open-ended reasoning~\cite{xu2026visionreward, he2024videoscore, he2025videoscore2, wang2026unifiedthink, wang2026unifiedflex, zhao2026envisioning}. However, they lack a sample-specific, explicitly verifiable evaluation contract specifying what to assess and the evidence supporting each judgment. This leads to three limitations:
\textit{(1) Incomplete inspection}: evaluators often focus on salient global properties while overlooking prompt-specific attributes, inter-object relations, brief actions, temporal ordering, and localized visual defects.
\textit{(2) Unfaithful justification}: unconstrained rationales may serve as post-hoc justifications for a predicted score rather than faithfully reflect the evidence and decisions underlying it.
\textit{(3) Entangled attribution}: when the same open-ended reasoning process is reused across evaluation dimensions, a single issue may be interpreted simultaneously as a semantic mismatch, a physical inconsistency, and a perceptual defect, resulting in duplicated penalties.

\begin{figure}[t]
    \centering
    \includegraphics[width=\columnwidth]{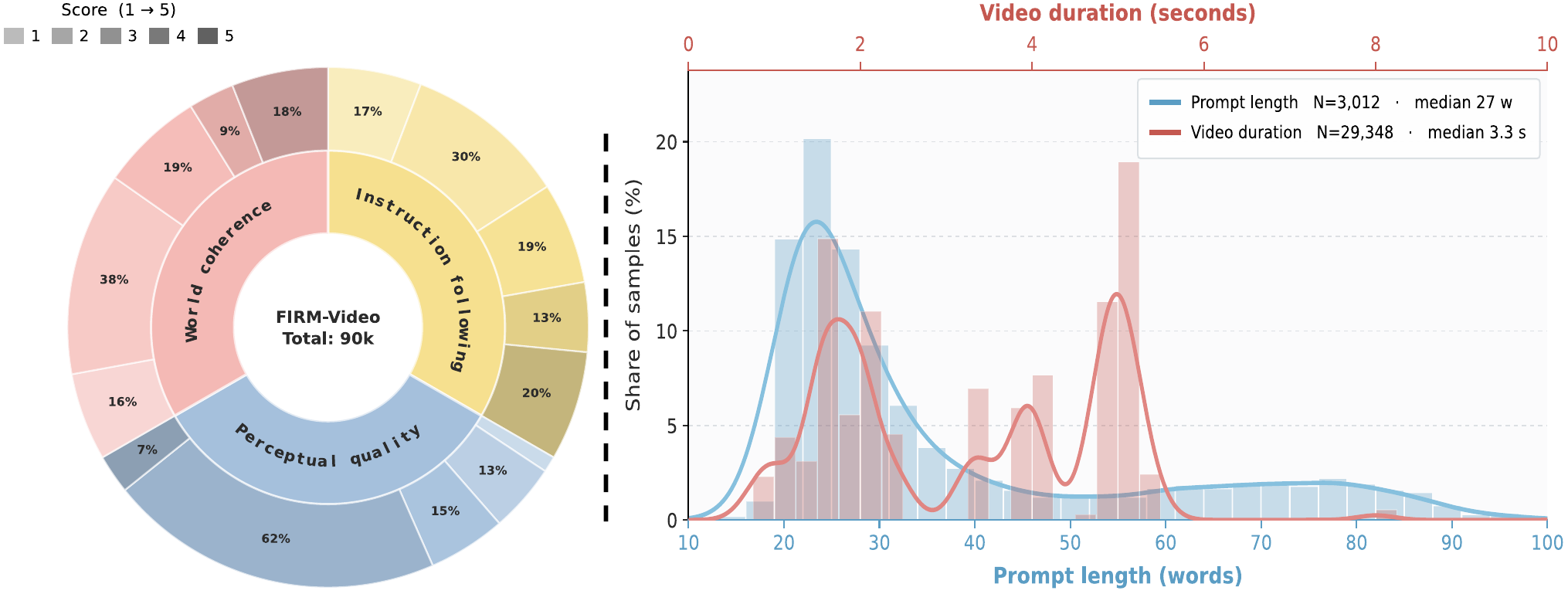}
    \caption{
        Statistics of the \textbf{FIRM-Video-90K} dataset. \textbf{Left:} distributions across evaluation dimensions and score ranges. \textbf{Right:} distributions of video duration and prompt length.
    }
    \label{fig:data_distribution}
\end{figure}

Therefore, we introduce \textbf{FIRM-Video}, a unified checklist-driven data construction framework for reliable T2V reward modeling under a \textbf{check-before-score} principle: construct dimension-specific checklists, verify each criterion against temporal visual evidence, and aggregate only the verified decisions. The checklist source reflects the nature of each dimension. For Instruction Following (IF), FIRM-Video decomposes the prompt into weighted atomic requirements and aggregates their verified satisfaction. For World Coherence (WC), it grounds visible entities and actions to construct target-specific checks, using the prompt to avoid penalizing intended fantasy or stylization and target-level aggregation to prevent excessive penalties in content-rich videos. For Perceptual Quality (PQ), it applies a generic taxonomy of visual defects, using the prompt only to distinguish intentional styles from rendering errors. Finally, the verified structured criteria and aggregated scores are transformed into natural-language analyses and ratings, yielding high-quality supervision for training end-to-end T2V reward models.

Applying this pipeline to a diverse collection of generated videos, we construct \textbf{FIRM-Video-90K}, comprising 88,044 dimension-specific reasoning-and-score instances derived from 29,348 videos and 3,012 unique prompts. We also introduce \textbf{FIRM-Video-Bench}, a benchmark consisting of 250 carefully selected videos evaluated by human experts, resulting in 750 fine-grained annotations across IF, PQ, and WC. It provides a unified and fair testbed for evaluating agreement between video reward models and human judgments.

Using FIRM-Video-90K, we train two reward models initialized from Qwen3-VL-8B~\cite{bai2025qwen3vl} and InternVL3-8B~\cite{zhu2025internvl3}, both substantially outperform their corresponding base models. In particular, FIRM-Video-8B based on Qwen3-VL reduces the overall MAE on FIRM-Video-Bench from 1.33 to 0.78, achieving the best overall MAE among all evaluated proprietary and open-source models, and remains competitive in preference accuracy on MJ-Bench-Video. We further evaluate its effectiveness through best-of-N sampling. With $N = 8$, FIRM-Video-8B achieves the highest VBench Total, Quality, and Semantic Scores across LaVie-Base~\cite{wang2025lavie}, CogVideoX-2B~\cite{yang2025cogvideox}, and Wan2.1-T2V-1.3B~\cite{wan2025wan}. Compared with the strongest competing selector, it improves the Total Score by 0.27–1.01 points and the Semantic Score by 1.24–2.11 points across the three generators. These results demonstrate that FIRM-Video-8B provides reliable absolute assessments and candidate rankings, establishing a practical foundation for evaluation, inference-time selection, and future reward-based alignment.

Our contributions are summarized as follows:
\begin{itemize}
  \item We propose FIRM-Video, a unified checklist-driven data construction framework for reliable T2V reward modeling that combines checklist construction, evidence verification, and score aggregation. Its data construction pipeline yields FIRM-Video-90K, containing 88,044 dimension-specific supervision instances over 29,348 videos.
  \item We construct FIRM-Video-Bench, a carefully expert-annotated point-wise benchmark of 250 videos with 750 dimension-specific annotations, covering instruction following, perceptual quality, and world coherence.
  \item We train two FIRM-Video-8B reward models from Qwen3-VL-8B and InternVL3-8B. Both substantially improve human alignment over their base models, while the Qwen3-VL-based model achieves the best overall MAE on FIRM-Video-Bench and consistently enhances Best-of-N selection across multiple T2V generators.
\end{itemize}

\section{Related Work}

\subsection{Text-to-Video Generation Models}

Text-to-video generation has progressed from early U-Net-based diffusion models~\cite{bar2024lumiere, ronneberger2015u} to larger DiT~\cite{peebles2023scalable} and flow-matching architectures~\cite{ma2024sit}, bringing clear gains in visual quality, temporal consistency, and instruction following. Early systems such as ModelScope~\cite{wang2023modelscope} and VideoCrafter2~\cite{chen2024videocrafter2} helped establish practical generation pipelines, while recent open models, including CogVideoX~\cite{yang2025cogvideox}, HunyuanVideo~\cite{kong2024hunyuanvideo}, Wan~\cite{wan2025wan}, and StepVideo-T2V~\cite{ma2025step}, have made high-quality video synthesis more accessible. Meanwhile, proprietary models such as Sora~\cite{OpenAIsora}, Kling~\cite{Kling-1.6}, Veo~\cite{Google-Veo3}, Pika~\cite{Pika2.2}, and Seedance~\cite{seedance2026seedance} continue to improve resolution, motion realism, long-range coherence, and controllability. Despite these advances, a systematic and human-aligned approach to evaluating video quality across different dimensions, from visual perception to semantic reasoning, is still lacking, highlighting the need for more comprehensive and interpretable evaluation frameworks.

\subsection{Reward Modeling for Visual Generation}

Reward modeling has become an important tool for aligning visual generation models with human preferences. Early studies~\cite{huang2024vbench, liu2024evalcrafter, huang2023t2i} mainly relied on fixed metrics such as FID~\cite{heusel2017gans} and CLIP~\cite{radford2021learning} Score to assess perceptual quality and text–visual alignment, but these metrics often show limited agreement with human judgments. Subsequent methods, including ImageReward~\cite{xu2023imagereward}, PickScore~\cite{kirstain2023pick}, and HPS~\cite{wu2023better}, learned reward functions from human annotations via point-wise regression or pair-wise preference learning, typically using Bradley–Terry or ranking objectives. 

Recent advances in large vision-language models have enabled VLM-based reward modeling. Early methods, including VideoReward~\cite{liu2026improving}, UnifiedReward~\cite{wang2025unified}, and Q-Insight~\cite{li2026q}, primarily predict scalar or multi-dimensional scores, while newer approaches adopt generative judging to produce assessments and rationales. VideoScore2~\cite{he2025videoscore2}, for example, generates detailed multi-dimensional evaluations but relies on costly human supervision and subjective, model-expanded explanations that are difficult to verify. UnifiedReward-Flex~\cite{wang2026unifiedflex} adapts criteria to each input, yet its pair-wise formulation yields only relative preferences rather than point-wise scores. FIRM~\cite{zhao2026trust} introduces task-specific critics for image generation and editing but does not address video-specific temporal quality. To address these limitations, we propose FIRM-Video, a checklist-driven data construction framework for reliable T2V reward modeling. Following a check-before-score paradigm, it constructs dimension-specific checklists, verifies each criterion against temporal visual evidence, and aggregates verified decisions. The resulting high-quality supervision enables end-to-end reward models to produce reliable, interpretable evaluations with strong agreement with human judgments.
\begin{figure*}[t]
    \centering
    \includegraphics[width=\textwidth]{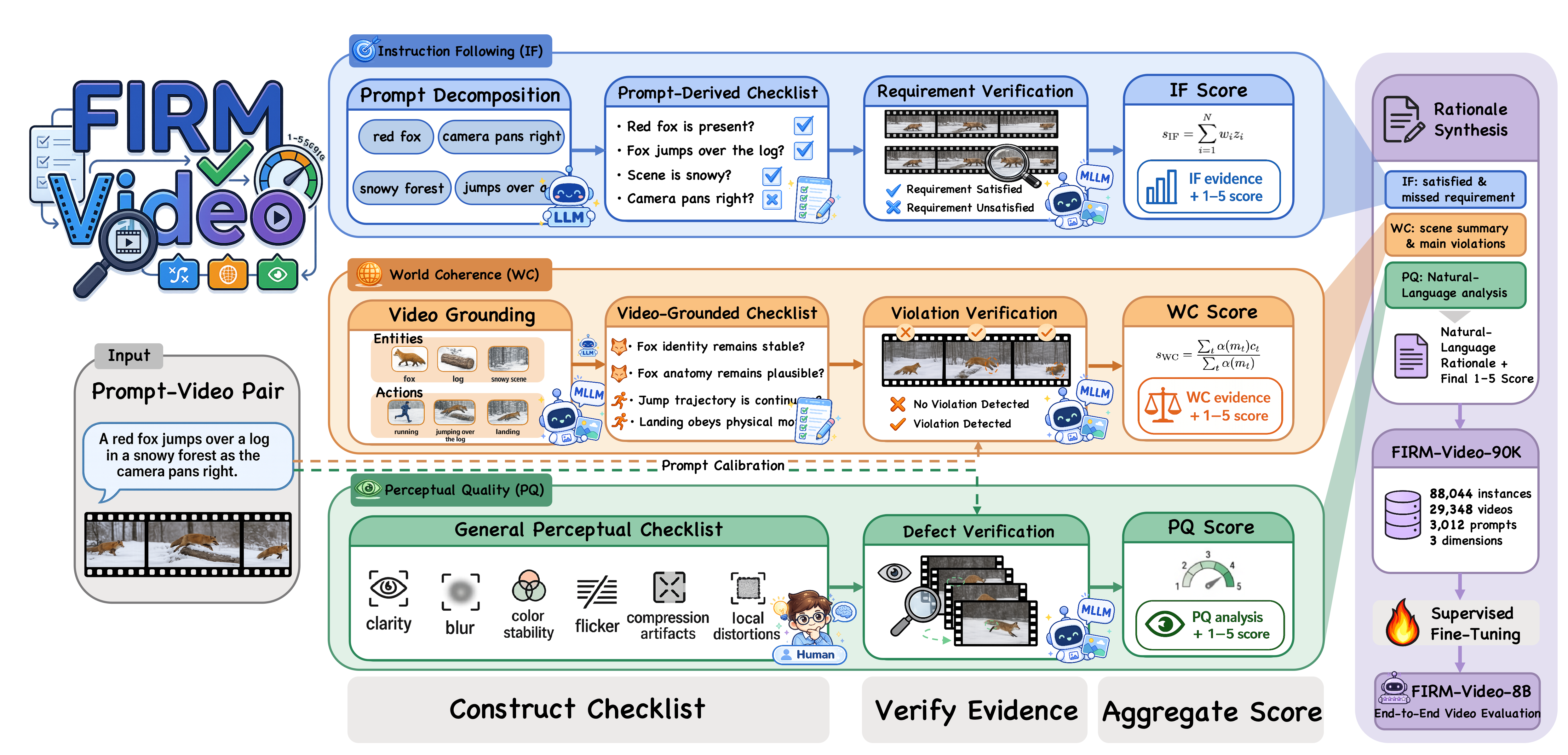}
    \caption{
    Overview of the FIRM-Video data construction framework and reward model training.
    Given a prompt and its generated video, FIRM-Video applies a unified
    \textit{check-before-score} pipeline to construct fine-grained supervision for
    instruction following, world coherence, and perceptual quality.
    The resulting dimension-specific analyses and scores form FIRM-Video-90K,
    which is then used to train the FIRM-Video-8B reward model.
    }
    \label{fig:firm_pipeline}
\end{figure*}

\section{Methodology}

\subsection{Overview of FIRM-Video}

\subsubsection{Data Collection}

We construct our dataset from two human preference datasets: Text2Video-Human Preferences, provided by Rapidata, and VideoFeedback2~\cite{he2025videoscore2}. After filtering, we retain approximately 3K text--video pairs from the former and 27K from the latter, resulting in about 30K samples. The dataset covers outputs generated by over 20 video generation models with diverse capability levels, including Sora2~\cite{OpenAIsora}, Veo3~\cite{Google-Veo3}, HunyuanVideo~\cite{kong2024hunyuanvideo}, Pika2.2~\cite{Pika2.2}, and Wan2.1~\cite{wan2025wan}. For each prompt, videos are typically sampled from around 10 randomly selected models. Their resolutions range from 256$\times$256 to 1980$\times$982, with frame rates between 8 and 30 fps.

\subsubsection{Dataset Statistics}

As shown in Figure~\ref{fig:data_distribution}, FIRM-Video-90K comprises approximately 90K fine-grained annotations for 29,348 videos generated from 3,012 prompts, with roughly 30K annotations for each evaluation dimension. Overall, the annotations cover the full range of score levels with relatively balanced distributions. Among the three dimensions, Instruction Following and World Coherence exhibit broad score distributions, whereas Perceptual Quality is skewed toward higher scores, reflecting that current video generation models generally produce visually acceptable videos with few obvious artifacts. In addition, prompts have a median length of 27 words, while videos have a median duration of 3.3 seconds, with most lasting between 1 and 6 seconds.

\subsection{Construction of FIRM-Video Data Pipeline}

FIRM-Video follows a unified \textit{check-before-score} paradigm. Given a prompt and a generated video, it first constructs dimension-specific checklists, verifies them against the video, and then aggregates the verification results into scores for Instruction Following (IF), World Coherence (WC), and Perceptual Quality (PQ), as illustrated in Figure~\ref{fig:firm_pipeline}. Specifically, IF derives an adaptive checklist from the prompt, WC constructs checklists from entities and actions grounded in the generated video, and PQ adopts a generic checklist of common perceptual defects carefully curated by human.

\subsubsection{Instruction Following}

Instruction Following evaluates how accurately and completely the generated video adheres to the input text prompt, including the requested subjects, attributes, actions, scenes, visual styles, camera viewpoints, and temporal ordering of events. 
The evaluation consists of two stages.

\textit{Stage 1: Prompt Decomposition.}
A text planner (Qwen3-32B) decomposes the prompt into a set of atomic yes--no questions covering all requested visual elements, while excluding audio-related requirements. Each question corresponds to a single observable requirement and is assigned an importance score $m_i\in\{1,\ldots,5\}$, where a higher value indicates greater importance. Videos generated from the same prompt share an identical checklist.

\textit{Stage 2: Checklist Verification.}
A proprietary multimodal evaluator takes the prompt, temporally ordered video frames, and the checklist as input. For each question, it predicts a binary answer (``yes'' or ``no'') together with brief visual evidence. Requirements that cannot be verified from the video are treated as unsatisfied.

Let $z_i\in\{0,1\}$ denote the verification result for question $i$. The IF score is computed as
\begin{equation}
s_{\mathrm{IF}}
= \sum_{i=1}^{N} w_i z_i,
\qquad
w_i = \frac{m_i}{\sum_{j=1}^{N}m_j}.
\label{eq}
\end{equation}
This score represents the importance-weighted proportion of prompt requirements satisfied by the generated video, and is subsequently mapped to a five-point rating.

\subsubsection{World Coherence}

World Coherence evaluates whether the generated content is consistent with everyday knowledge, common sense, and basic physical laws, with particular attention to temporal inconsistencies, unnatural or incoherent motion, physically implausible interactions, and structural abnormalities or deformations in objects and characters. 
Such violations may involve content not explicitly mentioned in the prompt, and directly constructing verification questions from the video can overlook peripheral entities or transient actions. We therefore adopt a three-stage pipeline.

\textit{Stage 1: Video Grounding.}
A multimodal grounding model (Qwen3-VL-235B-A22B-Instruct) analyzes the video without access to the prompt and produces a structured scene summary, an inventory of entities, and a list of actions or interactions. Each grounded target $t$ is assigned an importance score $m_t\in\{1,\ldots,5\}$. This explicit grounding improves target coverage while reducing prompt-induced bias.

\textit{Stage 2: Checklist Construction.}
A text planner (Qwen3-32B) converts each grounded target into a set of verification questions. For entities, the questions assess temporal consistency and structural plausibility; for actions, they evaluate motion quality and physical plausibility. Each question inherits the importance of its corresponding target, while incomplete target coverage is automatically detected and regenerated. The prompt is introduced only to distinguish intentional fantasy, stylization, or unusual motion from genuine violations: the grounding stage determines \textit{what} to evaluate, whereas the prompt specifies \textit{how} it should be interpreted.

\textit{Stage 3: Checklist Verification.}
A proprietary multimodal evaluator receives the video together with the generated questions. Neither the prompt nor target importance is provided at this stage. A ``yes'' response indicates a detected violation, whereas ``no'' indicates that no violation is observed.

Let $\mathcal{Q}_t$ denote the set of questions associated with target $t$, and let $b_j=1$ indicate that question $j$ detects a violation. We compute
\begin{equation}
s_{\mathrm{WC}}
= \frac{\sum_t \alpha(m_t)c_t}
{\sum_t \alpha(m_t)},
\qquad
c_t
= 1-\frac{\sum_{j\in\mathcal{Q}_t}b_j}
{|\mathcal{Q}_t|},
\label{eq2}
\end{equation}
where $\alpha(m_t)$ is a predefined increasing weight for importance level $m_t\in\{1,\ldots,5\}$. The target-level clean ratio provides partial credit when only some aspects of a target are affected, while avoiding unnecessary penalties for content-rich videos that naturally require more verification questions. The resulting score is then mapped to a five-point rating.

\subsubsection{Perceptual Quality}

Perceptual Quality assesses the perceptual and technical quality of the rendered video, including resolution, global and local clarity, sharpness or blurriness, brightness and color stability, flickering, compression artifacts, and other low-level visual distortions that may affect the viewing experience.
A proprietary multimodal evaluator analyzes the video using a generic, manually curated checklist of common perceptual defects and directly generates a checklist-based natural-language rationale with a five-point score. The prompt is provided only to distinguish intentional stylistic choices, such as unusual lighting, from unintended defects, rather than to assess semantic alignment.

\begin{table*}[t]
\centering

\setlength{\tabcolsep}{3pt}
\renewcommand{\arraystretch}{1.35}
\small

\resizebox{\textwidth}{!}{%
\begin{tabular}{
    l
    c c c c
    c c c c
    c c c c
    c c c
}
\toprule
\multirow{2}{*}{Model}
& \multicolumn{4}{c}{Instruction Following}
& \multicolumn{4}{c}{World Coherence}
& \multicolumn{4}{c}{Perceptual Quality}
& \multicolumn{3}{c}{Overall} \\

\cmidrule(lr){2-5}
\cmidrule(lr){6-9}
\cmidrule(lr){10-13}
\cmidrule(lr){14-16}

& MAE$\downarrow$
& STD$\downarrow$
& \makecell{Acc./Relaxed\\Acc.$\uparrow$}
& SRCC$\uparrow$

& MAE$\downarrow$
& STD$\downarrow$
& \makecell{Acc./Relaxed\\Acc.$\uparrow$}
& SRCC$\uparrow$

& MAE$\downarrow$
& STD$\downarrow$
& \makecell{Acc./Relaxed\\Acc.$\uparrow$}
& SRCC$\uparrow$

& MAE$\downarrow$
& STD$\downarrow$
& \makecell{Acc./Relaxed\\Acc.$\uparrow$} \\

\midrule

\multicolumn{16}{l}{\textbf{\textit{Closed-source models}}} \\
\addlinespace[1pt]

GPT-5
& \textbf{0.62}
& \textbf{0.71}
& \textbf{0.50/0.88}
& \textbf{0.74}
& 1.66
& 1.19
& 0.20/0.48
& 0.49
& 0.95
& 0.86
& 0.34/0.76
& \underline{0.51}
& 1.08
& 1.03
& 0.35/0.71 \\

Gemini-3.1-Pro
& 0.77
& 0.76
& 0.40/0.86
& \underline{0.71}
& 1.27
& 1.08
& 0.28/0.62
& \textbf{0.54}
& 1.44
& 1.12
& 0.21/0.59
& 0.41
& 1.16
& 1.04
& 0.30/0.69 \\

Doubao-Seed-2.0-Lite
& 0.80
& 0.83
& 0.41/0.84
& 0.67
& 1.56
& 1.24
& 0.24/0.53
& 0.44
& \textbf{0.80}
& \underline{0.75}
& \textbf{0.38/0.84}
& \textbf{0.55}
& 1.05
& 1.03
& 0.35/0.73 \\

\midrule

\multicolumn{16}{l}{\textbf{\textit{Open-source models}}} \\
\addlinespace[1pt]

InternVL3-8B
& 1.26
& 0.95
& 0.24/0.60
& 0.57
& 2.10
& 1.32
& 0.15/0.36
& 0.22
& 1.29
& 1.00
& 0.24/0.61
& 0.22
& 1.56
& 1.17
& 0.21/0.52 \\

InternVL3-38B
& 0.82
& 0.83
& 0.40/0.82
& 0.65
& 2.11
& 1.32
& 0.15/0.35
& 0.15
& 1.33
& 1.08
& 0.27/0.58
& 0.20
& 1.42
& 1.21
& 0.27/0.58 \\

Qwen3-VL-8B
& 0.93
& 0.89
& 0.34/0.80
& 0.60
& 1.79
& 1.35
& 0.22/0.46
& 0.29
& 1.28
& 1.04
& 0.28/0.60
& 0.34
& 1.33
& 1.16
& 0.28/0.62 \\

Qwen3-VL-30B-A3B
& 0.95
& 0.87
& 0.32/0.80
& 0.62
& 1.90
& 1.31
& 0.19/0.40
& 0.22
& 1.36
& 1.09
& 0.28/0.55
& 0.37
& 1.40
& 1.17
& 0.27/0.58 \\

Qwen3-VL-235B-A22B
& 0.93
& 0.87
& 0.33/0.80
& 0.65
& 1.63
& 1.30
& 0.24/0.52
& 0.31
& 1.27
& 1.04
& 0.28/0.60
& 0.40
& 1.27
& 1.12
& 0.29/0.64 \\

\midrule

\multicolumn{16}{l}{\textbf{\textit{Our methods}}} \\
\addlinespace[1pt]

\rowcolor{black!6}
FIRM-Video Data Pipeline$^\dagger$ 
& 0.68
& 0.68
& 0.43/0.90
& 0.69
& 0.80
& 0.83
& 0.41/0.84
& 0.63
& 0.74
& 0.70
& 0.40/0.86
& 0.61
& 0.73
& 0.74
& 0.41/0.87 \\

\addlinespace[2pt]

FIRM-Video-8B (InternVL3-8B)
& 0.69
& \underline{0.73}
& \underline{0.45/0.87}
& 0.67
& \underline{0.96}
& \underline{0.95}
& \underline{0.37/0.76}
& 0.49
& 0.90
& 0.77
& 0.32/0.80
& 0.35
& \underline{0.85}
& \underline{0.83}
& \underline{0.38/0.81} \\

FIRM-Video-8B (Qwen3-VL-8B)
& \underline{0.65}
& 0.77
& \textbf{0.50/0.88}
& 0.69
& \textbf{0.86}
& \textbf{0.88}
& \textbf{0.40/0.80}
& \underline{0.53}
& \underline{0.82}
& \textbf{0.74}
& \underline{0.36/0.83}
& \underline{0.51}
& \textbf{0.78}
& \textbf{0.80}
& \textbf{0.42/0.84} \\

\bottomrule
\end{tabular}%
}

\caption{
Performance comparison on \textbf{FIRM-Video-Bench} across three
evaluation dimensions: Instruction Following (IF), World Coherence (WC),
and Perceptual Quality (PQ).
The two values under Acc./Relaxed Acc. denote strict and relaxed
accuracy, respectively.
The row marked with $\dagger$ reports the performance of direct data construction pipeline and is excluded from ranking.}
\label{tab:firm-video-bench}

\end{table*}

\subsection{Synthesis of FIRM-Video-90K Dataset }

The explicit \textit{check-before-score} pipelines provide reliable supervision by decomposing evaluation into criterion-level verification followed by evidence-based score aggregation. However, executing these multi-stage pipelines with multiple specialized models for every video is computationally expensive. We therefore run the pipelines offline to synthesize high-quality supervision for end-to-end reward modeling. Specifically, criterion-level verification results are transformed into natural-language rationales and paired with their corresponding evidence-derived scores.

The three evaluation pipelines produce complementary verification results. IF identifies satisfied and unmet prompt requirements, WC provides scene grounding together with target-level coherence violations, and PQ evaluates perceptual quality using a generic checklist. We transform these structured outputs into concise, dimension-specific natural-language rationales. For IF and WC, a LLM (Qwen3-32B) performs this transformation. The IF rationale begins with a summary of satisfied prompt requirements, followed by an explanation of unmet or incorrectly fulfilled ones. The WC rationale begins with a brief scene summary, followed by the major detected violations organized by target and evaluation aspect. For PQ, the evaluator already generates a coherent checklist-based rationale, which is used directly. In all cases, the score is determined before rationale generation, ensuring that the natural-language analysis verbalizes the verified decisions without altering the final judgment.

We retain only videos with valid evaluations across IF, WC, and PQ, yielding one training instance per dimension for each video. The input includes the original prompt, sampled video frames, and the corresponding evaluation instruction. The target is organized as a structured JSON object containing the synthesized natural-language rationale and the associated five-point score. After filtering, 29,348 videos yield 88,044 training instances, forming \textbf{FIRM-Video-90K}.

FIRM-Video-90K is then used to train FIRM-Video-8B, which directly predicts dimension-specific rationales and scores from the prompt and video frames. During inference, the model performs evaluation in a single forward pass without explicitly constructing checklists or invoking the planner, grounding model, and verifier used during data synthesis. Consequently, the proposed framework distills the reliability of explicit checklist-based evaluation into an efficient end-to-end reward model, substantially reducing inference cost while preserving evaluation quality.

\subsection{Construction of FIRM-Video-Bench}

\subsubsection{Data Collection and Annotation}

FIRM-Video-Bench consists of 250 prompt--video pairs sampled from VideoFeedback2 and VideoPhy2~\cite{bansal2025videophy}, resulting in a total of 750 dimension-specific evaluation annotations across IF, WC, and PQ. We retain only the prompt--video pairs and re-annotate every sample according to our unified evaluation definitions. The original annotations are discarded because they were collected under different evaluation criteria and contain annotation noise. FIRM-Video-Bench shares no overlapping videos with FIRM-Video-90K.

Each sample is independently annotated by multiple human experts using integer scores from 1 to 5 for each evaluation dimension, with disagreements resolved through discussion. Annotators are provided with the prompt, the complete video, and the same sampled frames used by the reward models. The complete video enables holistic assessment, while the sampled frames ensure consistency between human judgments and the visual evidence available to the models.

\subsubsection{Data Balance and Evaluation Metrics}

FIRM-Video-Bench is balanced across all rating levels to cover a broad range of generation quality. Every score level is represented in each evaluation dimension, and no single level accounts for more than $30.4\%$ of the samples. This balanced distribution reduces bias toward majority-score prediction and better evaluates fine-grained scoring performance.

We report Mean Absolute Error (MAE), exact accuracy, relaxed accuracy, and Spearman's rank correlation coefficient (SRCC) as the primary evaluation metrics. Exact accuracy measures exact score agreement with human annotations, while relaxed accuracy counts predictions within one rating level of the ground-truth score as correct. Together with MAE and SRCC, these metrics evaluate prediction error, discrete agreement, and ranking consistency.

\section{Experiments}

\subsection{Experimental Setting}

\subsubsection{Implementation Details}

We initialize FIRM-Video-8B with Qwen3-VL-8B~\cite{bai2025qwen3vl} and InternVL3-8B~\cite{zhu2025internvl3}, respectively, and fully fine-tune both models on FIRM-Video-90K. During training and inference, eight frames are uniformly sampled from each video. We use a batch size of 2 with 32 gradient accumulation steps, an initial learning rate of $5 \times 10^{-6}$, cosine decay, and a warm-up ratio of 0.05. Models are trained for two epochs using LLaMA-Factory framework~\cite{zheng2024llamafactory}.

\subsubsection{Evaluation Protocol}

We primarily evaluate agreement with human judgments on FIRM-Video-Bench and assess out-of-domain generalization on MJ-Bench-Video~\cite{tong2025mj}. Following VideoScore2~\cite{he2025videoscore2}, we evaluate three dimensions that align with the FIRM-Video evaluation taxonomy—Alignment, Consistency \& Coherence, and Fitness—and additionally include an Overall dimension for evaluation on MJ-Bench-Video. We report pair-wise preference accuracy across all four dimensions. We further evaluate the effectiveness of FIRM-Video-8B through best-of-N sampling on VBench~\cite{huang2024vbench}, where each method selects the highest-scoring video from $N$ candidates.

\begin{figure}[t]
\centering

\begin{minipage}[t]{0.49\textwidth}
    \centering
    \vspace{0pt}

    \begingroup
    \small
    \setlength{\tabcolsep}{2.6pt}
    \renewcommand{\arraystretch}{1.36}

    \resizebox{\linewidth}{!}{%
    \begin{tabular}{@{}lcccc@{}}
    \toprule
    \textbf{Model}
    & \textbf{Alignment} $\uparrow$
    & \textbf{C\&C} $\uparrow$
    & \textbf{Fitness} $\uparrow$
    & \textbf{Overall} $\uparrow$ \\
    \midrule

    \multicolumn{5}{@{}l}{%
    \textbf{\textit{Closed-source models}}} \\
    \addlinespace[1pt]

    GPT-5
    & 29.52
    & 24.54
    & 19.29
    & 26.09 \\

    Gemini-3.1-Pro
    & \underline{31.53}
    & 28.72
    & 28.24
    & \textbf{33.06} \\

    Doubao-Seed-2.0-Lite
    & 30.43
    & 26.69
    & 24.70
    & 26.21 \\

    \addlinespace[2pt]
    \midrule
    \addlinespace[1pt]

    \multicolumn{5}{@{}l}{%
    \textbf{\textit{Open-source models}}} \\
    \addlinespace[1pt]

    InternVL3-8B
    & 28.41
    & 24.58
    & 22.96
    & 27.21 \\

    InternVL3-38B
    & 28.17
    & 22.04
    & 23.70
    & 26.93 \\

    Qwen3-VL-8B
    & 29.80
    & 22.37
    & 27.32
    & 21.17 \\

    Qwen3-VL-30B-A3B
    & 28.96
    & 25.60
    & \textbf{35.06}
    & 22.24 \\

    Qwen3-VL-235B-A22B
    & 30.03
    & 21.86
    & 29.17
    & 23.40 \\

    \addlinespace[2pt]
    \midrule
    \addlinespace[1pt]

    \multicolumn{5}{@{}l}{%
    \textbf{\textit{Ours}}} \\
    \addlinespace[1pt]

    FIRM-Video-8B (Qwen3-VL-8B)
    & 31.09
    & \underline{30.41}
    & \underline{32.10}
    & 27.81 \\

    FIRM-Video-8B (InternVL3-8B)
    & \textbf{31.93}
    & \textbf{35.35}
    & 29.41
    & \underline{31.06} \\

    \bottomrule
    \end{tabular}%
    }

    \captionof{table}{
        Preference accuracy on MJ-Bench-Video across Alignment,
        C\&C, Fitness, and Overall. Best results are
        \textbf{bolded} and second-best results are
        \underline{underlined}.
    }
    \label{tab:mj-bench-video}

    \endgroup
\end{minipage}
\hfill
\begin{minipage}[t]{0.49\textwidth}
    \centering
    \vspace{0pt}

    \includegraphics[
        width=\linewidth
    ]{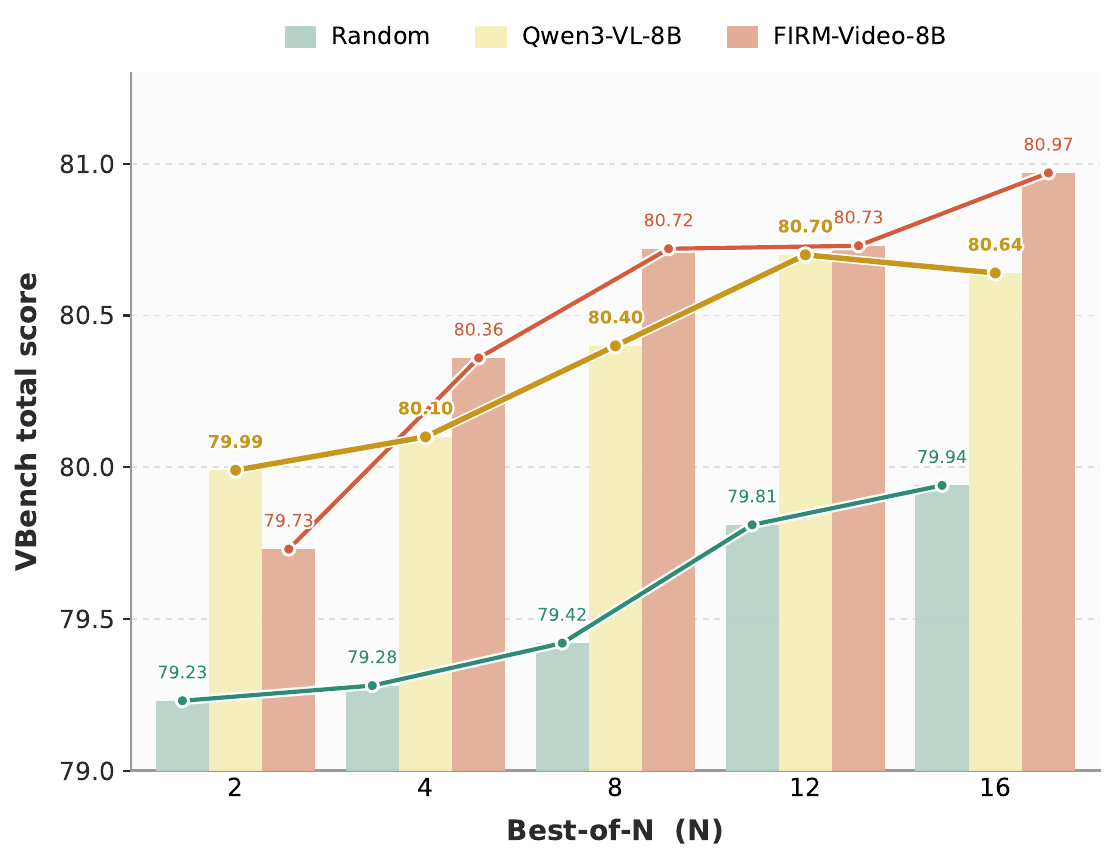}

    \captionof{figure}{
        Best-of-N scaling results on VBench using LaVie-Base
        as the video generation model. Performance is measured
        by the VBench total score as $N$ increases.
    }
    \label{fig:best_of_n_scaling}
\end{minipage}

\end{figure}

\subsection{Results on FIRM-Video-Bench}

We compare FIRM-Video-8B with proprietary models, including GPT-5~\cite{gpt-5}, Gemini-3.1-Pro~\cite{gemini-3.1-pro}, and Doubao-Seed-2.0-Lite~\cite{seed2026seed2}, and open-source models from the Qwen3-VL~\cite{bai2025qwen3vl} and InternVL3~\cite{zhu2025internvl3} families. We report mean absolute error (MAE), the standard deviation of absolute errors (STD), point-wise accuracy, relaxed accuracy within one point, and Spearman's rank correlation coefficient (SRCC).

As shown in Table~\ref{tab:firm-video-bench}, proprietary models generally outperform existing open-source baselines, especially on IF and WC. GPT-5 obtains the lowest IF MAE of 0.62, while Doubao-Seed-2.0-Lite achieves the best PQ MAE of 0.80. Both FIRM-Video-8B variants substantially improve over their respective base models. In particular, FIRM-Video-8B (Qwen3-VL-8B) achieves the best overall MAE of 0.78, together with strict and relaxed accuracies of 0.42 and 0.84, outperforming all proprietary and open-source baselines. It also obtains the lowest WC MAE while remaining highly competitive on IF and PQ. We also report the performance of our data construction pipeline as a direct-scoring ablation, confirming the effectiveness of check-before-score.

\subsection{Results on MJ-Bench-Video}

MJ-Bench-Video is a video preference benchmark comprising 2,170 videos annotated by human across five dimensions. We evaluate the three dimensions aligned with the FIRM-Video taxonomy: Alignment, Consistency \& Coherence (C\&C), Fitness, and additionally report Overall preference using the average score across the three dimensions. We report pair-wise preference accuracy with ties, where preferences are determined by comparing the scores assigned to each video pair. For Overall, pairs with an average score difference within 0.5 are treated as ties. Correct predictions receive a score of 1 for preference pairs and 0.5 for tie pairs.

As shown in Table~\ref{tab:mj-bench-video}, FIRM-Video demonstrates strong out-of-domain generalization, achieving competitive performance against both proprietary models and substantially larger open-source alternatives. FIRM-Video-8B (InternVL3-8B) achieves the best results on Alignment and C\&C while ranking second on Overall, whereas FIRM-Video-8B (Qwen3-VL-8B) attains the second-best performance on both C\&C and Fitness. These results further demonstrate FIRM-Video-8B's strong capability for effective preference-based candidate ranking.

\begin{table*}[t]
\centering

\begingroup
\small

\setlength{\tabcolsep}{3pt}
\renewcommand{\arraystretch}{1.35}
\setlength{\aboverulesep}{0.30ex}
\setlength{\belowrulesep}{0.30ex}

\resizebox{\textwidth}{!}{%
\begin{tabular}{l*{11}{c}}
\toprule

\multirow[c]{2}{*}[-1.6ex]{
    \makecell[c]{\textbf{Sampling}\\\textbf{Strategy}}
}
& \multirow[c]{2}{*}[-1.6ex]{
    \makecell[c]{\textbf{Total}\\\textbf{Score}}
}
& \multirow[c]{2}{*}[-1.6ex]{
    \makecell[c]{\textbf{Quality}\\\textbf{Score}}
}
& \multirow[c]{2}{*}[-1.6ex]{
    \makecell[c]{\textbf{Semantic}\\\textbf{Score}}
}
& \multicolumn{4}{c}{\textbf{Video Quality}}
& \multicolumn{4}{c}{\textbf{Video--Condition Consistency}} \\

\cmidrule(lr){5-8}
\cmidrule(lr){9-12}

& & &

& \makecell[c]{Subject\\Consistency}
& \makecell[c]{Background\\Consistency}
& \makecell[c]{Temporal\\Flickering}
& \makecell[c]{Imaging\\Quality}

& \makecell[c]{Multiple\\Objects}
& \makecell[c]{Color}
& \makecell[c]{Spatial\\Relationship}
& \makecell[c]{Appearance\\Style} \\

\midrule

\multicolumn{12}{l}{\textbf{\textit{T2V model: LaVie-Base}}} \\

Random
& 79.42
& 81.34
& 71.77
& 92.09
& 97.64
& 97.63
& 65.92
& 37.73
& \textbf{89.65}
& 43.30
& 23.94 \\

By Qwen3-VL-8B
& 80.03
& 81.61
& 73.73
& 92.64
& 97.56
& 97.60
& 66.88
& 50.91
& 83.11
& 47.32
& \underline{24.27} \\

By InternVL3-8B
& \underline{80.45}
& \underline{82.10}
& \underline{73.82}
& 92.94
& 97.56
& 97.56
& \underline{67.02}
& 50.30
& 85.85
& \underline{45.81}
& 24.10 \\

By VideoScore2
& 79.95
& 81.61
& 73.33
& \underline{92.95}
& \underline{97.80}
& \underline{97.76}
& 66.54
& \underline{53.73}
& 86.86
& 45.70
& 23.87 \\

\rowcolor{black!6}
By FIRM-Video-8B
& \textbf{80.72}
& \textbf{82.14}
& \textbf{75.06}
& \textbf{93.66}
& \textbf{97.91}
& \textbf{98.12}
& \textbf{67.46}
& \textbf{58.99}
& \underline{88.20}
& \textbf{51.20}
& \textbf{24.40} \\

\midrule

\multicolumn{12}{l}{\textbf{\textit{T2V model: CogVideoX-2B}}} \\

Random
& 78.57
& \underline{80.90}
& 69.25
& 93.47
& 96.66
& 96.90
& 59.72
& 49.16
& 88.43
& 61.30
& 22.95 \\

By Qwen3-VL-8B
& 78.49
& 80.22
& 71.59
& 94.48
& 96.84
& 96.90
& 59.20
& 55.56
& 88.40
& \underline{65.80}
& \textbf{23.30} \\

By InternVL3-8B
& 78.35
& 80.27
& 70.65
& 94.53
& \underline{96.89}
& \underline{97.04}
& 59.90
& 51.30
& \underline{89.49}
& 61.09
& \underline{23.24} \\

By VideoScore2
& \underline{78.67}
& 80.30
& \underline{72.17}
& \underline{94.63}
& 96.78
& 96.98
& \textbf{60.08}
& \underline{61.05}
& 87.59
& 63.82
& 22.54 \\

\rowcolor{black!6}
By FIRM-Video-8B
& \textbf{79.68}
& \textbf{81.03}
& \textbf{74.28}
& \textbf{94.92}
& \textbf{97.28}
& \textbf{97.33}
& \underline{60.00}
& \textbf{62.88}
& \textbf{92.37}
& \textbf{66.86}
& 23.02 \\

\midrule

\multicolumn{12}{l}{\textbf{\textit{T2V model: Wan2.1-T2V-1.3B}}} \\

Random
& 80.40
& 84.13
& 65.49
& 94.57
& 98.08
& 98.98
& \underline{68.18}
& 57.01
& 85.63
& 76.13
& 19.73 \\

By Qwen3-VL-8B
& 81.43
& 84.30
& 69.96
& 94.57
& 97.90
& 98.87
& 67.57
& 67.91
& 85.94
& 74.39
& \textbf{20.67} \\

By InternVL3-8B
& \underline{81.70}
& 84.43
& \underline{70.78}
& \underline{95.46}
& \underline{98.32}
& 98.84
& 67.71
& 67.61
& \underline{86.15}
& \underline{77.08}
& 20.30 \\

By VideoScore2
& 81.64
& \underline{84.86}
& 68.74
& 95.45
& 98.13
& \underline{98.99}
& 67.57
& \underline{72.26}
& 83.91
& 74.56
& 19.98 \\

\rowcolor{black!6}
By FIRM-Video-8B
& \textbf{82.36}
& \textbf{84.91}
& \textbf{72.19}
& \textbf{96.54}
& \textbf{98.68}
& \textbf{99.22}
& \textbf{68.62}
& \textbf{73.40}
& \textbf{92.23}
& \textbf{81.10}
& \underline{20.64} \\

\bottomrule
\end{tabular}%
}

\endgroup

\caption{
Best-of-N performance comparison of FIRM-Video-8B and other sampling
strategies on VBench. 
The best and second-best results under each T2V generator are highlighted in \textbf{bold} and \underline{underlined}, respectively.
}
\label{tab:vbench-bon}

\end{table*}

\subsection{Best-of-N Sampling on VBench}

We conduct best-of-N sampling with $N=8$ on three text-to-video generators of varying capabilities: LaVie-Base~\cite{wang2025lavie}, CogVideoX-2B~\cite{yang2025cogvideox}, and Wan2.1-T2V-1.3B~\cite{wan2025wan}. For each prompt, eight candidates are ranked by FIRM-Video-8B (Qwen3-VL-8B), Qwen3-VL-8B, InternVL3-8B, or VideoScore2, with random sampling as the baseline.

VBench comprises 16 disentangled dimensions, evenly split between Video Quality and Video–Condition Consistency. Their weighted aggregates form the Quality Score and Semantic Score, which are further combined into the Total Score. We report these three aggregate metrics and four representative dimensions from each group. Since VBench evaluates each dimension independently, we adopt dimension-aware sampling rather than simply aggregating the three FIRM-Video scores into a single overall score. Specifically, each VBench dimension is ranked using the most relevant FIRM-Video evaluation score. This design also enables fine-grained ablations of the contribution of each FIRM-Video evaluation dimension. The complete results and dimension-mapping rules are provided in the supplementary material.

As shown in Table~\ref{tab:vbench-bon}, FIRM-Video-8B achieves the best Total, Quality, and Semantic Scores across all three generators. Compared with the strongest competing method, it improves the Total Score by 0.27, 1.01, and 0.66, respectively, and the Semantic Score by 1.24--2.11. It also ranks first on most representative dimensions, including subject consistency, multiple objects, and spatial relationship. These consistent gains across generators demonstrate its robustness for perceptual-quality and semantic-alignment assessment.

\subsection{Scaling with the Number of Samples}

We further evaluate best-of-N scaling on VBench using LaVie-Base, varying $N$ over ${2,4,8,12,16}$. Random sampling, Qwen3-VL-8B, and FIRM-Video-8B (Qwen3-VL-8B) select one video from each candidate set, which is then evaluated using the VBench Total Score.

As shown in Figure~\ref{fig:best_of_n_scaling}, performance generally improves with larger candidate pools. Although FIRM-Video-8B is slightly below Qwen3-VL-8B at $N=2$, it performs best for all $N \geq 4$, with its score increasing from 80.36 to 80.97. At $N=16$, it surpasses random sampling and Qwen3-VL-8B by 1.03 and 0.33 points, respectively, indicating more reliable ranking as the candidate pool grows.

\section{Limitations}

This work mainly validates FIRM-Video through human-aligned evaluation and best-of-N sampling. While these results demonstrate the models' ability to provide reliable reward
signals, we have not yet used them to directly optimize T2V generators. Future work may integrate FIRM-Video-8B with reinforcement learning or preference optimization to improve instruction following, world coherence, and perceptual quality during text-to-video generation.



\section{Conclusion}

We introduced FIRM-Video, a checklist-driven data construction framework for reliable T2V reward modeling under the \textit{check-before-score} principle. It constructs dimension-specific checklists for Instruction Following, World Coherence, and Perceptual Quality, verifies each criterion against temporal visual evidence, and aggregates only verified decisions into traceable scores. This framework yielded FIRM-Video-90K and the expert-annotated FIRM-Video-Bench. Reward models trained on FIRM-Video-90K substantially improved agreement with human judgments over their base models, with the Qwen3-VL-based FIRM-Video-8B achieving the best overall MAE on FIRM-Video-Bench and the strongest best-of-N sampling performance across three T2V generators on VBench. These results demonstrate that checklist-driven supervision enables both accurate absolute assessment and effective candidate ranking, advancing reliable evaluation and alignment of T2V models. Future work will explore its use in reinforcement learning and preference optimization for improving T2V generation.

\bibliographystyle{plainnat}
\bibliography{main}

\clearpage
\beginappendix
\setcounter{section}{0}
\numberwithin{equation}{section}
\section{Implementation Details of the Data Pipeline}
\label{app:pipeline}

\subsection{Instruction-Following Score Computation}
\label{app:if-score}

For each prompt, the planner decomposes the instruction into $N$ atomic visual requirements, each assigned an importance score $m_i\in\{1,\ldots,5\}$. Videos generated from the same prompt share the same checklist, and audio-only requirements are excluded.

The normalized weight of requirement $i$ is
\begin{equation}
w_i=\frac{m_i}{\sum_{j=1}^{N}m_j}.
\label{eq:app-if-weight}
\end{equation}

For serialization, weights are rounded to two decimal places. Positive weights rounded to zero are set to $0.01$, and the residual is assigned to the largest weight to ensure $\sum_i w_i=1$.

Let $z_i\in\{0,1\}$ denote the verifier output for requirement $i$, where unverifiable requirements are treated as unsatisfied. The IF score is
\begin{equation}
s_{\mathrm{IF}}
=\sum_{i=1}^{N} w_i z_i,
\qquad
s_{\mathrm{IF}}\in[0,1].
\label{eq:app-if-score}
\end{equation}

The final score is rounded to four decimal places and mapped to a five-point rating as described in Appendix~\ref{app:score-mapping}.

\subsection{World-Coherence Score Computation}
\label{app:wc-score}

Video grounding produces a set of entity and action targets, each assigned an importance score $m_t\in\{1,\ldots,5\}$. Each target is converted into a set of verification questions $\mathcal{Q}_t$, and every question inherits the importance of its corresponding target.

Let $b_j=1$ denote that question $j$ detects a violation. The clean ratio of target $t$ is
\begin{equation}
c_t
=1-\frac{1}{|\mathcal{Q}_t|}
\sum_{j\in\mathcal{Q}_t}b_j.
\label{eq:app-wc-clean}
\end{equation}

Importance is converted to weights using
\begin{equation}
\alpha(m)=
\begin{cases}
1.0,&m=1,\\
1.5,&m=2,\\
2.5,&m=3,\\
3.5,&m=4,\\
5.0,&m=5.
\end{cases}
\label{eq:app-wc-importance}
\end{equation}

The WC score is computed as
\begin{equation}
s_{\mathrm{WC}}
=\frac{\sum_t\alpha(m_t)c_t}
{\sum_t\alpha(m_t)},
\qquad
s_{\mathrm{WC}}\in[0,1].
\label{eq:app-wc-score}
\end{equation}

As with IF, the resulting score is rounded to four decimal places before conversion to the five-point rating (Appendix~\ref{app:score-mapping}).

\subsection{Score Mapping}
\label{app:score-mapping}

Continuous checklist scores are mapped to five-point ratings using empirically determined thresholds. Let $s \in [0,1]$ denote the normalized checklist score and $R(s)$ the corresponding rating.

For the \textbf{IF} dimension,
\[
R_{\mathrm{IF}}(s)=
\begin{cases}
1, & 0 \le s < 0.35,\\
2, & 0.35 \le s < 0.70,\\
3, & 0.70 \le s < 0.85,\\
4, & 0.85 \le s < 1.00,\\
5, & s = 1.00,
\end{cases}
\]
where a rating of 5 requires all positively weighted requirements to be satisfied.

For the \textbf{WC} dimension,
\[
R_{\mathrm{WC}}(s)=
\begin{cases}
1, & 0 \le s < 0.20,\\
2, & 0.20 \le s < 0.58,\\
3, & 0.58 \le s < 0.78,\\
4, & 0.78 \le s < 0.89,\\
5, & 0.89 \le s \le 1.00.
\end{cases}
\]

\subsection{Ablation of Score Aggregation}
\label{app:aggregation-ablation}

We compare importance-weighted aggregation with uniform averaging while keeping the checklists, verification results, and score mapping fixed.

\textit{Instruction Following.}
The uniform baseline assigns equal weight to all requirements:
\begin{equation}
    s_{\mathrm{IF}}^{\mathrm{mean}}
    =\frac{1}{N}\sum_{i=1}^{N} z_i,
    \label{eq:app-aggregation-if-mean}
\end{equation}
whereas our method uses the importance-weighted score in Eq.~\ref{eq:app-if-score}.

\textit{World Coherence.}
The uniform baseline averages the clean ratios of all $T$ targets:
\begin{equation}
    s_{\mathrm{WC}}^{\mathrm{mean}}
    =\frac{1}{T}\sum_{t} c_t,
    \label{eq:app-aggregation-wc-mean}
\end{equation}
whereas our method uses the importance-weighted score in Eq.~\ref{eq:app-wc-score}.


\begin{table}[!htbp]
\centering
\begingroup
\small
\setlength{\tabcolsep}{5pt}
\renewcommand{\arraystretch}{1.08}
\begin{tabular}{@{}ccccccc@{}}
\toprule
\textbf{Dimension} & \textbf{Aggregation} & \textbf{MAE} $\downarrow$ & \textbf{STD} $\downarrow$ & \textbf{Acc.} $\uparrow$ & \textbf{Relaxed Acc.} $\uparrow$ & \textbf{SRCC} $\uparrow$ \\
\midrule
\multirow{2}{*}{IF}
& mean & \textbf{0.68} & 0.70 & \textbf{0.44} & 0.88 & \textbf{0.69} \\
& \textbf{Importance weighted (ours)} & \textbf{0.68} & \textbf{0.68} & 0.43 & \textbf{0.90} & \textbf{0.69} \\
\midrule
\multirow{2}{*}{WC}
& mean & 0.86 & \textbf{0.83} & 0.37 & 0.82 & 0.61 \\
& \textbf{Importance weighted (ours)} & \textbf{0.80} & \textbf{0.83} & \textbf{0.41} & \textbf{0.84} & \textbf{0.63} \\
\bottomrule
\end{tabular}
\endgroup
\caption{Aggregation ablations on FIRM-Video-Bench. STD is computed over absolute errors; relaxed accuracy allows a one-point deviation from the expert rating. Best results within each dimension are bold.}
\label{tab:aggregation-ablation}
\end{table}

As shown in Table~\ref{tab:aggregation-ablation}, importance weighting improves IF stability and relaxed accuracy while preserving MAE and SRCC. For WC, it reduces MAE from 0.86 to 0.80 and improves accuracy, relaxed accuracy, and SRCC.

\section{Statistical Analysis of FIRM-Video-90K and FIRM-Video-Bench}
\label{app:data-statistics}

Table~\ref{tab:score-distributions} summarizes the statistics of FIRM-Video-90K and FIRM-Video-Bench.

\begin{table}[!htbp]
\centering
\begingroup
\small
\setlength{\tabcolsep}{4pt}
\renewcommand{\arraystretch}{1.15}
\resizebox{0.96\textwidth}{!}{%
\begin{tabular}{@{}cccccccc@{}}
\toprule
\textbf{Dataset} & \textbf{Dimension} & \textbf{Score=1} & \textbf{Score=2} & \textbf{Score=3} & \textbf{Score=4} & \textbf{Score 5} & \textbf{Total} \\
\midrule
\multirow{4}{*}{FIRM-Video-90K}
& IF & 5,104 (17.4\%) & 8,900 (30.3\%) & 5,551 (18.9\%) & 3,806 (13.0\%) & 5,987 (20.4\%) & 29,348 \\
& WC & 4,795 (16.3\%) & 11,129 (37.9\%) & 5,649 (19.2\%) & 2,510 (8.6\%) & 5,265 (17.9\%) & 29,348 \\
& PQ & 874 (3.0\%) & 3,749 (12.8\%) & 4,257 (14.5\%) & 18,331 (62.5\%) & 2,137 (7.3\%) & 29,348 \\
& \textbf{All} & \textbf{10,773} & \textbf{23,778} & \textbf{15,457} & \textbf{24,647} & \textbf{13,389} & \textbf{88,044} \\
\midrule
\multirow{4}{*}{FIRM-Video-Bench}
& IF & 28 (11.2\%) & 73 (29.2\%) & 63 (25.2\%) & 59 (23.6\%) & 27 (10.8\%) & 250 \\
& WC & 48 (19.2\%) & 76 (30.4\%) & 48 (19.2\%) & 44 (17.6\%) & 34 (13.6\%) & 250 \\
& PQ & 11 (4.4\%) & 54 (21.6\%) & 69 (27.6\%) & 59 (23.6\%) & 57 (22.8\%) & 250 \\
& \textbf{All} & \textbf{87} & \textbf{203} & \textbf{180} & \textbf{162} & \textbf{118} & \textbf{750} \\
\bottomrule
\end{tabular}%
}
\endgroup
\caption{Score distributions of FIRM-Video-90K and FIRM-Video-Bench.}
\label{tab:score-distributions}
\end{table}

\section{Best-of-N Sampling and Results on VBench}
\label{app:vbench-bon}

\subsection{Dimension Mapping}
\label{app:vbench-dimension-mapping}

The dimension mapping rule on VBench is shown in Table~\ref{tab:dim-mapping}. The candidate with the highest corresponding reward is selected for each dimension.

\begin{table}[!htbp]
\centering

\begin{tabular*}{\linewidth}{@{\extracolsep{\fill}}cll@{}}
\toprule
\textbf{Reward} & \textbf{VBench Dimension} & \textbf{VBench Group} \\
\midrule

\multirow{10}{*}{\textbf{IF}}
 & Dynamic Degree         & Quality \\
 & Overall Consistency    & Cond. Consist. \\
 & Object Class           & Cond. Consist. \\
 & Multiple Objects       & Cond. Consist. \\
 & Human Action           & Cond. Consist. \\
 & Color                  & Cond. Consist. \\
 & Spatial Relationship   & Cond. Consist. \\
 & Scene                  & Cond. Consist. \\
 & Appearance Style       & Cond. Consist. \\
 & Temporal Style         & Cond. Consist. \\
\addlinespace
\cmidrule(lr){1-3}

\multirow{3}{*}{\textbf{WC}}
 & Subject Consistency    & Quality \\
 & Background Consistency & Quality \\
 & Motion Smoothness      & Quality \\
\addlinespace
\cmidrule(lr){1-3}

\multirow{3}{*}{\textbf{PQ}}
 & Temporal Flickering    & Quality \\
 & Aesthetic Quality      & Quality \\
 & Imaging Quality        & Quality \\
\bottomrule
\end{tabular*}

\caption{Dimension mapping from VBench dimensions to FIRM-Video reward dimensions. ``Quality'' and ``Cond. Consist.'' denote VBench's Video-Quality and Video--Condition Consistency groups, respectively.}
\label{tab:dim-mapping}
\end{table}

\subsection{Detailed Results}
\label{app:vbench-detailed-results}

Detailed results on all 16 VBench subdimensions are reported in Tables~\ref{tab:vbench-bon-quality} and~\ref{tab:vbench-bon-condition}.

\begin{table}[!htbp]
\centering
\begingroup
\small
\setlength{\tabcolsep}{3pt}
\renewcommand{\arraystretch}{1.35}
\setlength{\aboverulesep}{0.30ex}
\setlength{\belowrulesep}{0.30ex}
\resizebox{\textwidth}{!}{%
\begin{tabular}{@{}cc*{7}{c}@{}}
\toprule
\textbf{T2V Model} & \textbf{Sampling Strategy} & \makecell{\textbf{Subject}\\\textbf{Consistency}} & \makecell{\textbf{Background}\\\textbf{Consistency}} & \makecell{\textbf{Temporal}\\\textbf{Flickering}} & \makecell{\textbf{Motion}\\\textbf{Smoothness}} & \makecell{\textbf{Dynamic}\\\textbf{Degree}} & \makecell{\textbf{Aesthetic}\\\textbf{Quality}} & \makecell{\textbf{Imaging}\\\textbf{Quality}} \\
\midrule
        \multirow{5}{*}{LaVie-Base}
        & Random          & 92.09 & 97.64 & 97.63 & \underline{96.63} & 55.56 & 64.53 & 65.92 \\
        & Qwen3-VL-8B    & 92.64 & 97.56 & 97.60 & 96.62 & 55.56 & \textbf{64.93} & 66.88 \\
        & InternVL3-8B    & 92.94 & 97.56 & 97.56 & \textbf{96.69} & \textbf{61.11} & \underline{64.74} & \underline{67.02} \\
        & VideoScore2     & \underline{92.95} & \underline{97.80} & \underline{97.76} & 96.39 & \underline{56.94} & 64.27 & 66.54 \\
        & \textbf{FIRM-Video-8B} & \textbf{93.66} & \textbf{97.91} & \textbf{98.12} & 96.58 & \underline{56.94} & 64.17 & \textbf{67.46} \\
        \midrule
        \multirow{5}{*}{CogVideoX-2B}
        & Random          & 93.47 & 96.66 & 96.90 & 97.23 & \textbf{73.61} & 58.47 & 59.72 \\
        & Qwen3-VL-8B    & 94.48 & 96.84 & 96.90 & \textbf{97.35} & 62.50 & 58.30 & 59.20 \\
        & InternVL3-8B    & 94.53 & \underline{96.89} & \underline{97.04} & 97.24 & 62.50 & 57.84 & 59.90 \\
        & VideoScore2     & \underline{94.63} & 96.78 & 96.98 & 97.27 & 59.72 & \textbf{59.31} & \textbf{60.08} \\
        & \textbf{FIRM-Video-8B} & \textbf{94.92} & \textbf{97.28} & \textbf{97.33} & \underline{97.34} & \underline{65.28} & \underline{59.15} & \underline{60.00} \\
        \midrule
        \multirow{5}{*}{Wan2.1-T2V-1.3B}
        & Random          & 94.57 & 98.08 & 98.98 & 98.21 & 62.50 & 64.38 & \underline{68.18} \\
        & Qwen3-VL-8B    & 94.57 & 97.90 & 98.87 & 98.10 & \underline{66.67} & 64.94 & 67.57 \\
        & InternVL3-8B    & \underline{95.46} & \underline{98.32} & 98.84 & 98.29 & 63.89 & 64.87 & 67.71 \\
        & VideoScore2     & 95.45 & 98.13 & \underline{98.99} & \underline{98.43} & \textbf{68.06} & \underline{65.10} & 67.57 \\
        & \textbf{FIRM-Video-8B} & \textbf{96.54} & \textbf{98.68} & \textbf{99.22} & \textbf{98.49} & 59.72 & \textbf{65.67} & \textbf{68.62} \\
\bottomrule
\end{tabular}%
}
\endgroup
\caption{Best-of-8 results on the video-quality subdimensions of VBench.}
\label{tab:vbench-bon-quality}
\end{table}

\begin{table}[!htbp]
\centering
\begingroup
\small
\setlength{\tabcolsep}{2.5pt}
\renewcommand{\arraystretch}{1.35}
\setlength{\aboverulesep}{0.30ex}
\setlength{\belowrulesep}{0.30ex}
\resizebox{\textwidth}{!}{%
\begin{tabular}{@{}cc*{9}{c}@{}}
\toprule
\textbf{T2V Model} & \textbf{Sampling Strategy} & \makecell{\textbf{Object}\\\textbf{Class}} & \makecell{\textbf{Multiple}\\\textbf{Objects}} & \makecell{\textbf{Human}\\\textbf{Action}} & \textbf{Color} & \makecell{\textbf{Spatial}\\\textbf{Relationship}} & \textbf{Scene} & \makecell{\textbf{Appearance}\\\textbf{Style}} & \makecell{\textbf{Temporal}\\\textbf{Style}} & \makecell{\textbf{Overall}\\\textbf{Consistency}} \\
\midrule
        \multirow{5}{*}{LaVie-Base}
        & Random          & \underline{93.12} & 37.73 & 92.00 & \textbf{89.65} & 43.30 & \underline{52.33} & 23.94 & 24.73 & 27.20 \\
        & Qwen3-VL-8B    & 91.61 & 50.91 & \textbf{96.00} & 83.11 & \underline{47.32} & \textbf{52.69} & \underline{24.27} & \textbf{25.25} & \textbf{27.75} \\
        & InternVL3-8B    & \textbf{95.81} & 50.30 & \textbf{96.00} & 85.85 & 45.81 & 51.02 & 24.10 & 25.03 & 27.46 \\
        & VideoScore2     & 91.06 & \underline{53.73} & \underline{93.00} & 86.86 & 45.70 & 50.80 & 23.87 & 25.16 & 27.33 \\
        & \textbf{FIRM-Video-8B} & 90.51 & \textbf{58.99} & \underline{93.00} & \underline{88.20} & \textbf{51.20} & 52.03 & \textbf{24.40} & \underline{25.19} & \underline{27.57} \\
        \midrule
        \multirow{5}{*}{CogVideoX-2B}
        & Random          & 76.11 & 49.16 & 87.00 & 88.43 & 61.30 & 39.90 & 22.95 & 23.77 & 24.43 \\
        & Qwen3-VL-8B    & 83.62 & 55.56 & \textbf{91.00} & 88.40 & \underline{65.80} & 35.83 & \textbf{23.30} & \underline{23.84} & \underline{25.27} \\
        & InternVL3-8B    & 82.91 & 51.30 & 89.00 & \underline{89.49} & 61.09 & 37.79 & \underline{23.24} & 23.74 & \textbf{25.30} \\
        & VideoScore2     & \underline{84.41} & \underline{61.05} & \textbf{91.00} & 87.59 & 63.82 & \underline{40.48} & 22.54 & 23.54 & 25.06 \\
        & \textbf{FIRM-Video-8B} & \textbf{85.21} & \textbf{62.88} & \underline{90.00} & \textbf{92.37} & \textbf{66.86} & \textbf{45.13} & 23.02 & \textbf{24.29} & 25.13 \\
        \midrule
        \multirow{5}{*}{Wan2.1-T2V-1.3B}
        & Random          & 74.76 & 57.01 & 68.00 & 85.63 & 76.13 & 24.35 & 19.73 & \underline{23.74} & 23.31 \\
        & Qwen3-VL-8B    & \underline{82.36} & 67.91 & \textbf{83.00} & 85.94 & 74.39 & 25.51 & \textbf{20.67} & 23.53 & \textbf{24.78} \\
        & InternVL3-8B    & \textbf{84.34} & 67.61 & \underline{82.00} & \underline{86.15} & \underline{77.08} & \textbf{30.38} & 20.30 & \textbf{23.86} & 24.15 \\
        & VideoScore2     & 79.27 & \underline{72.26} & 79.00 & 83.91 & 74.56 & 23.84 & 19.98 & 23.73 & 23.88 \\
        & \textbf{FIRM-Video-8B} & 80.22 & \textbf{73.40} & \textbf{83.00} & \textbf{92.23} & \textbf{81.10} & \underline{29.14} & \underline{20.64} & 23.52 & \underline{24.57} \\
\bottomrule
\end{tabular}%
}
\endgroup
\caption{Best-of-8 results on the video--condition consistency subdimensions of VBench.}
\label{tab:vbench-bon-condition}
\end{table}

\section{Inference Prompts}
\label{app:inference-prompts}

Here we present the inference prompts for FIRM-Video-8B and the compared baseline models. \texttt{\$\{n\_frames\}} and \texttt{\$\{t2v\_prompt\}} are replaced by the number of sampled frames and the T2V prompt.

\begin{center}
\begin{AIbox}{Instruction Following}
You are an expert for evaluating how well an AI-generated video follows its text prompt.

The video is given as ${n_frames} frames uniformly sampled in temporal order
(frame 1 = earliest, frame ${n_frames} = latest). Judge only from these
${n_frames} frames.

We evaluate the dimension 'instruction_following': how accurately and completely
the video follows the text prompt, including requested subjects, attributes,
actions, scene, style, camera/viewpoint, and temporal event order.

Text prompt:

${t2v_prompt}

Based on the video content, the text prompt, and the dimension definition, please
evaluate the video's instruction_following and give the score. The score must be
an integer in the range of 1 - 5. FIRST give a brief reasoning, THEN give the
score.

Your output must be a valid JSON object and nothing else, in exactly the
following format:
{
  "reasoning": "<reasoning>",
  "score": <score>
}

DO NOT include anything before or after the JSON object.
\end{AIbox}
\end{center}

\begin{center}
\begin{AIbox}{Perceptual Quality}
You are an expert for evaluating the perceptual quality of AI-generated videos.

The video is given as ${n_frames} frames uniformly sampled in temporal order
(frame 1 = earliest, frame ${n_frames} = latest). Judge only from these
${n_frames} frames.

We evaluate the dimension 'visual_quality': the perceptual and technical quality
of the rendered video, including resolution, overall and local clarity, sharpness
or blurriness, brightness and color stability, flicker, compression artifacts,
and low-level visual distortions that affect the viewing experience.

Text prompt (context only):

${t2v_prompt}

The text prompt is provided so that you can recognize intentional creative
choices it requests (e.g. a stylized look, a non-realistic scene, an unusual
camera) and NOT penalize them on this dimension. DO NOT judge how faithfully the
video follows the prompt here -- that is covered by a separate dimension.

Based on the video content and the dimension definition, please evaluate the
video's visual_quality and give the score. The score must be an integer in the
range of 1 - 5. FIRST give a brief reasoning, THEN give the score.

Your output must be a valid JSON object and nothing else, in exactly the
following format:
{
  "reasoning": "<reasoning>",
  "score": <score>
}

DO NOT include anything before or after the JSON object.
\end{AIbox}
\end{center}

\begin{center}
\begin{AIbox}{World Coherence}
You are an expert for evaluating whether an AI-generated video is coherent with
the real world.

The video is given as ${n_frames} frames uniformly sampled in temporal order
(frame 1 = earliest, frame ${n_frames} = latest). Judge only from these
${n_frames} frames.

We evaluate the dimension 'world_consistency': whether the video contains any
violations of everyday knowledge, common sense, or basic physical laws, including
temporal inconsistencies, unnatural or incoherent motion, physically implausible
interactions, and structural abnormalities or deformities in objects or
characters.

Text prompt (context only):

${t2v_prompt}

The text prompt is provided so that you can recognize intentional creative
choices it requests (e.g. a non-realistic scene or a deliberately fantastical
setup) and NOT penalize them on this dimension. DO NOT judge how faithfully the
video follows the prompt here -- that is covered by a separate dimension.

Based on the video content and the dimension definition, please evaluate the
video's world_consistency and give the score. The score must be an integer in
the range of 1 - 5. FIRST give a brief reasoning, THEN give the score.

Your output must be a valid JSON object and nothing else, in exactly the
following format:
{
  "reasoning": "<reasoning>",
  "score": <score>
}

DO NOT include anything before or after the JSON object.
\end{AIbox}
\end{center}

\section{Human Annotation Interface}
\label{app:human-annotation}

\begin{figure}[!htbp]
    \centering
    \includegraphics[width=\textwidth]{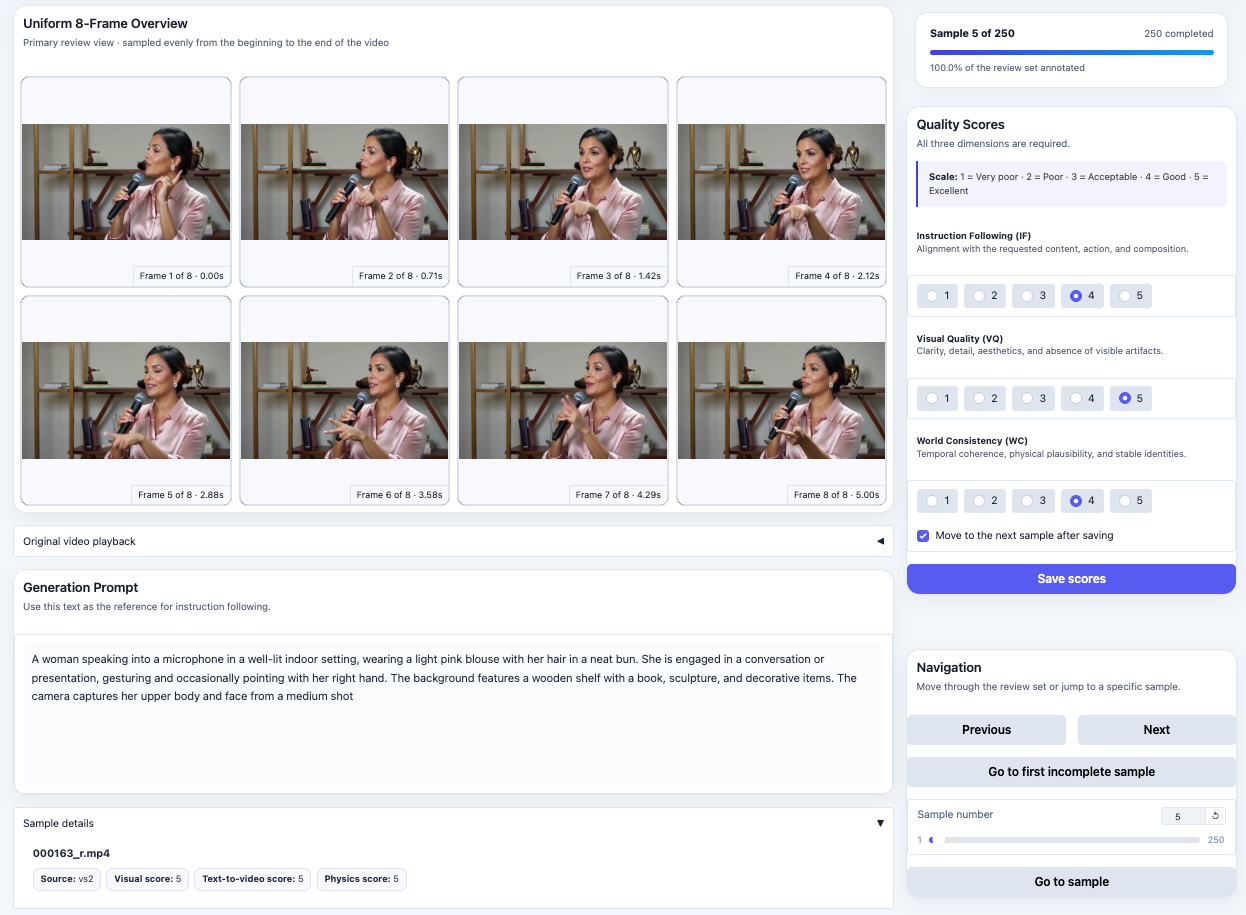}
    \caption{Human annotation interface for FIRM-Video-Bench. Annotators rate instruction following, perceptual quality, and world coherence on a five-point scale.}
    \label{fig:annotation-interface}
\end{figure}

Figure~\ref{fig:annotation-interface} shows the annotation interface used for human evaluation. Annotators rate IF, PQ, and WC independently on a five-point scale after reviewing the video, an eight-frame overview, and the generation prompt.

\section{Qualitative Examples}
\label{qualitative-results}

\begin{figure}[!htbp]
    \centering
    \includegraphics[width=\textwidth]{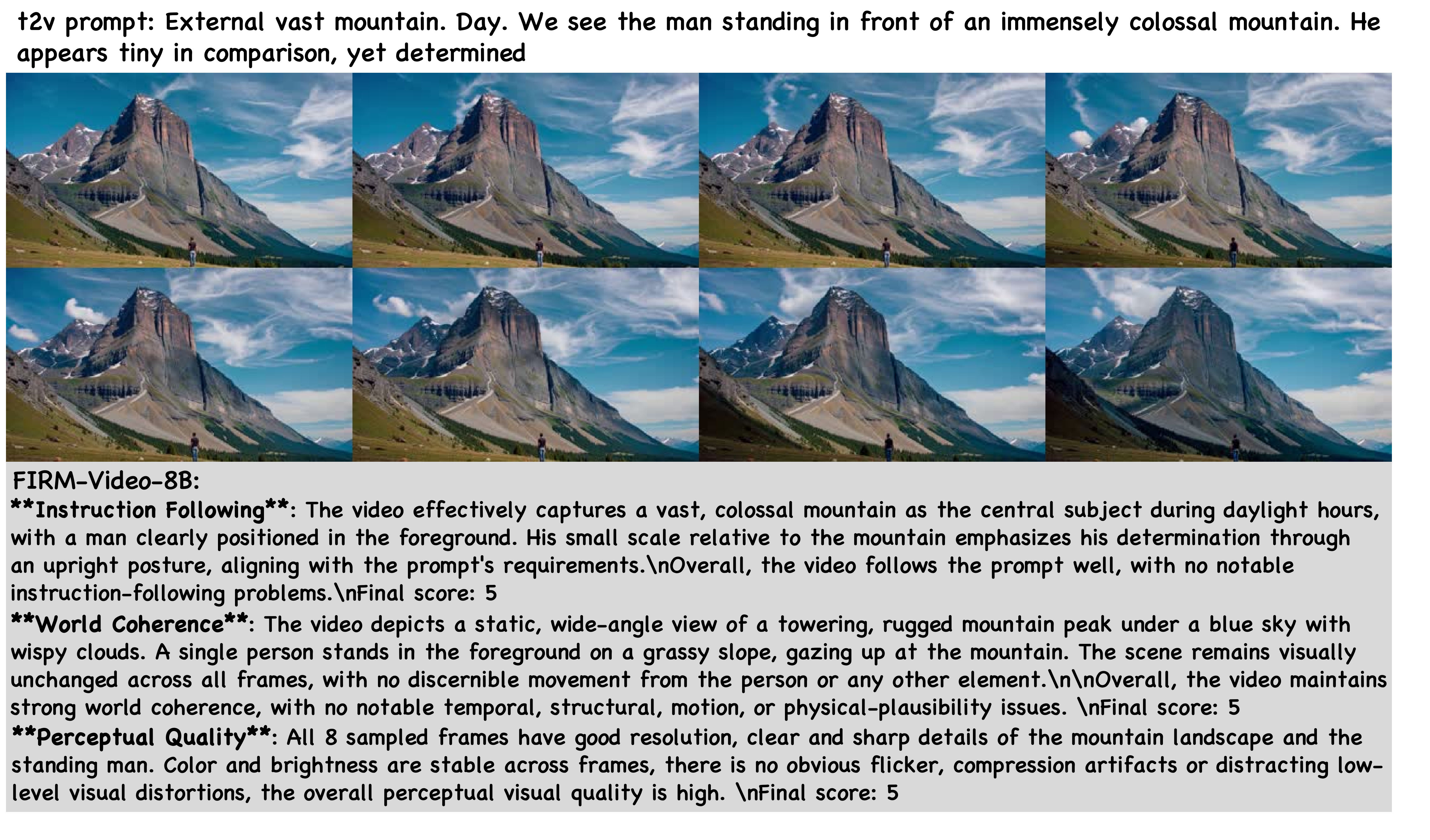}
    \caption{Qualitative example of FIRM-Video-8B evaluation (1)}
    \label{fig:case1}
\end{figure}

\begin{figure}[!htbp]
    \centering
    \includegraphics[width=\textwidth]{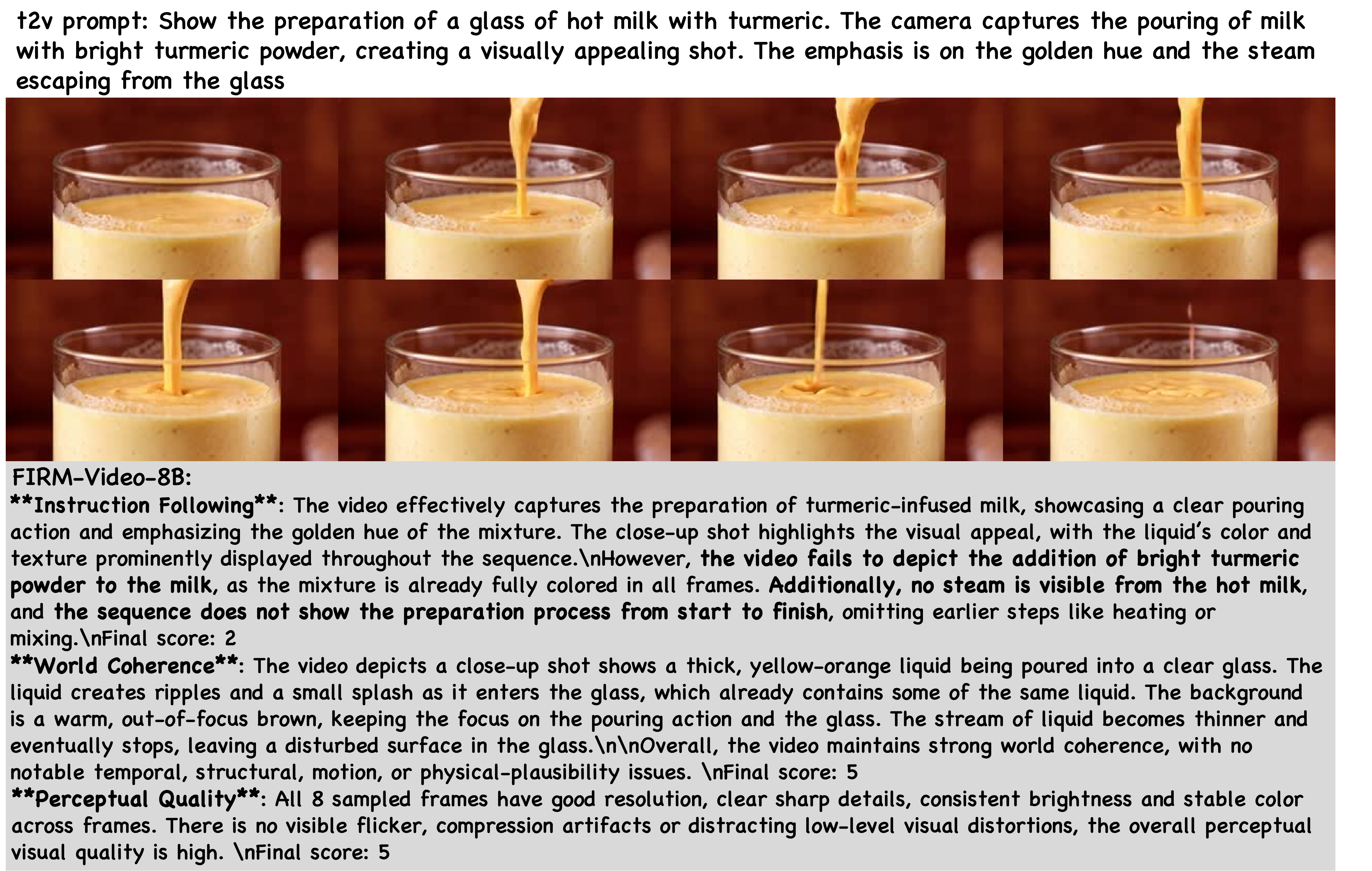}
    \caption{Qualitative example of FIRM-Video-8B evaluation (2)}
    \label{fig:case2}
\end{figure}

\begin{figure}[!htbp]
    \centering
    \includegraphics[width=\textwidth]{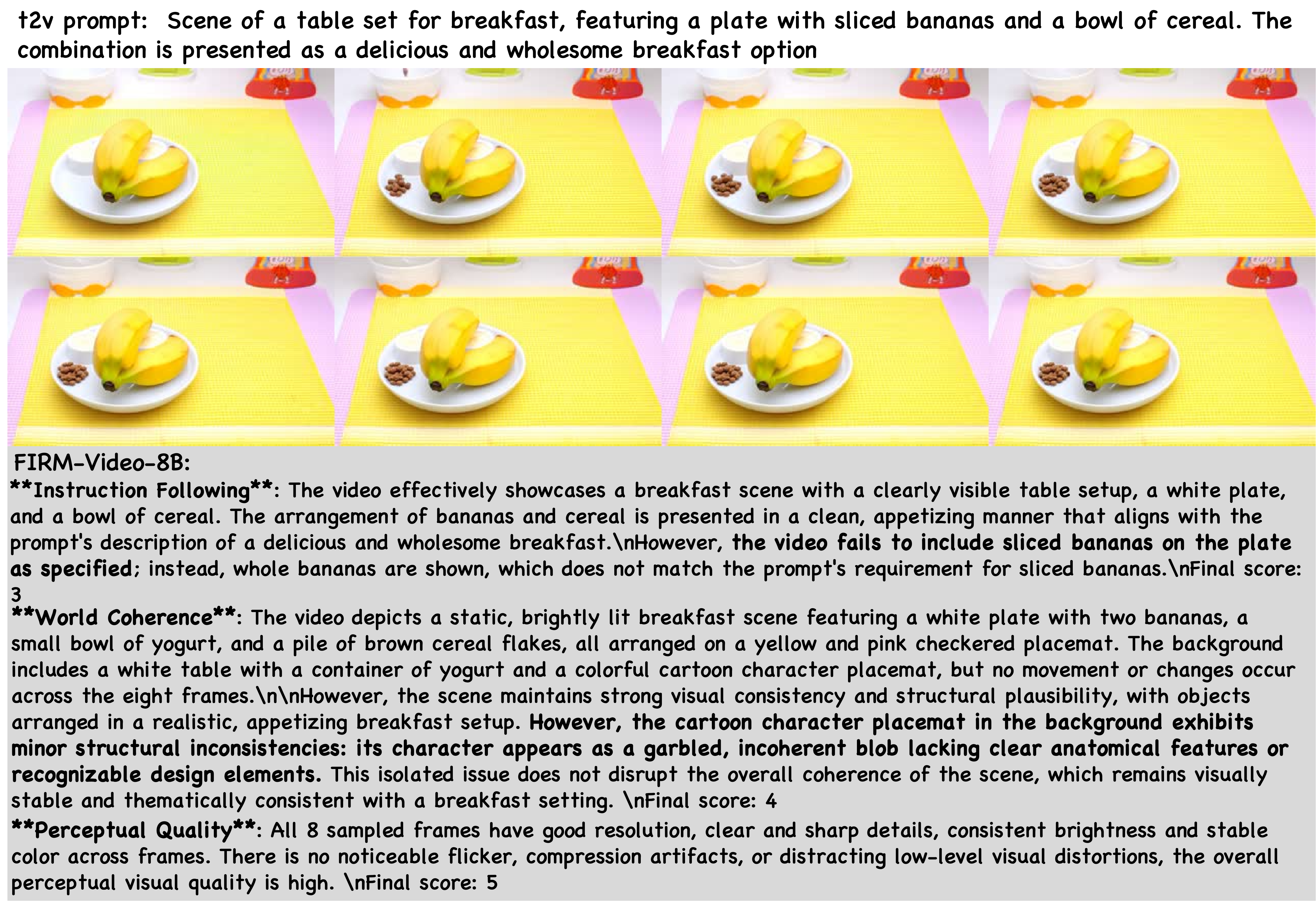}
    \caption{Qualitative example of FIRM-Video-8B evaluation (3)}
    \label{fig:case3}
\end{figure}

\begin{figure}[!htbp]
    \centering
    \includegraphics[width=\textwidth]{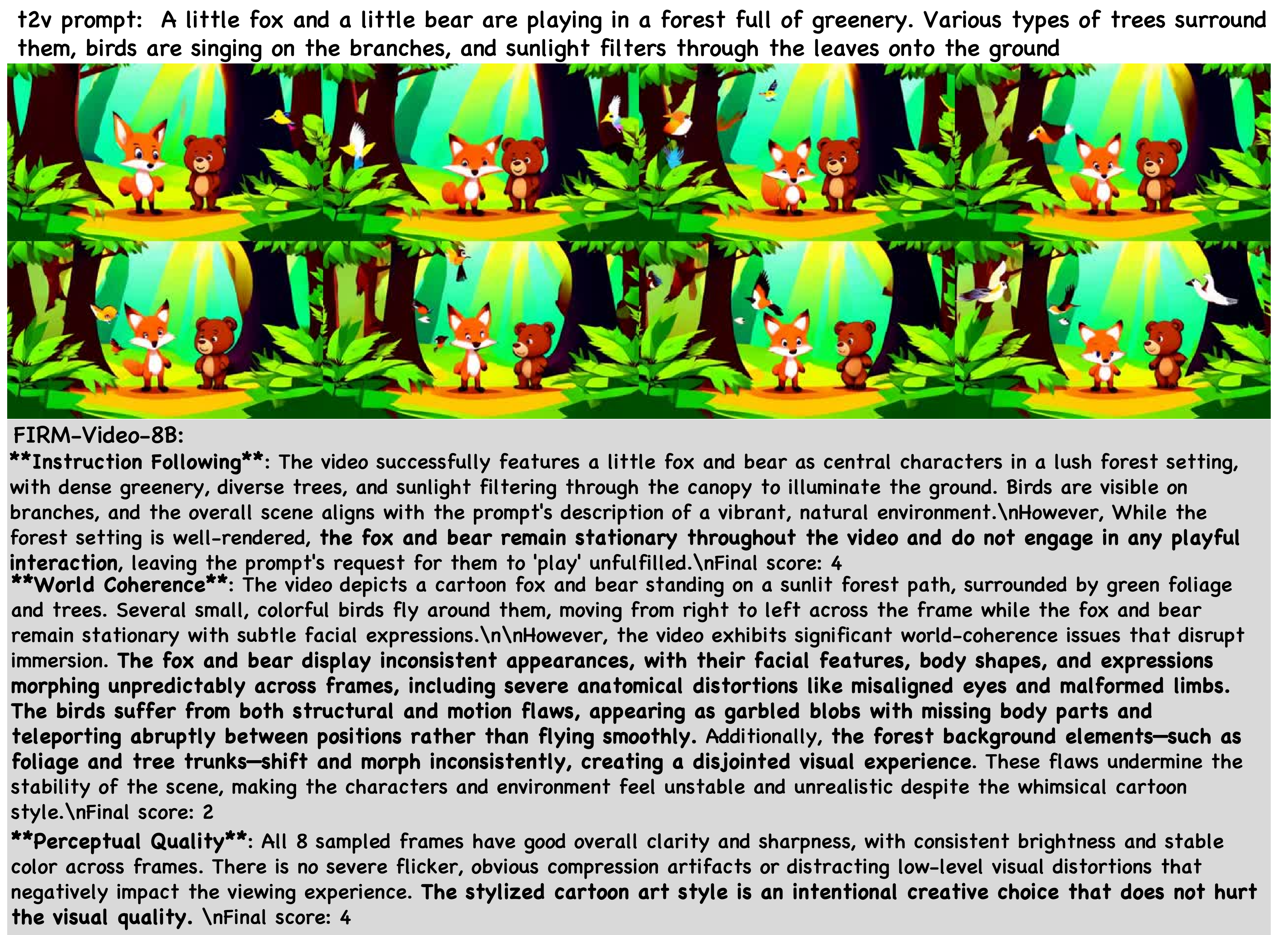}
    \caption{Qualitative example of FIRM-Video-8B evaluation (4)}
    \label{fig:case4}
\end{figure}

\begin{figure}[!htbp]
    \centering
    \includegraphics[width=\textwidth]{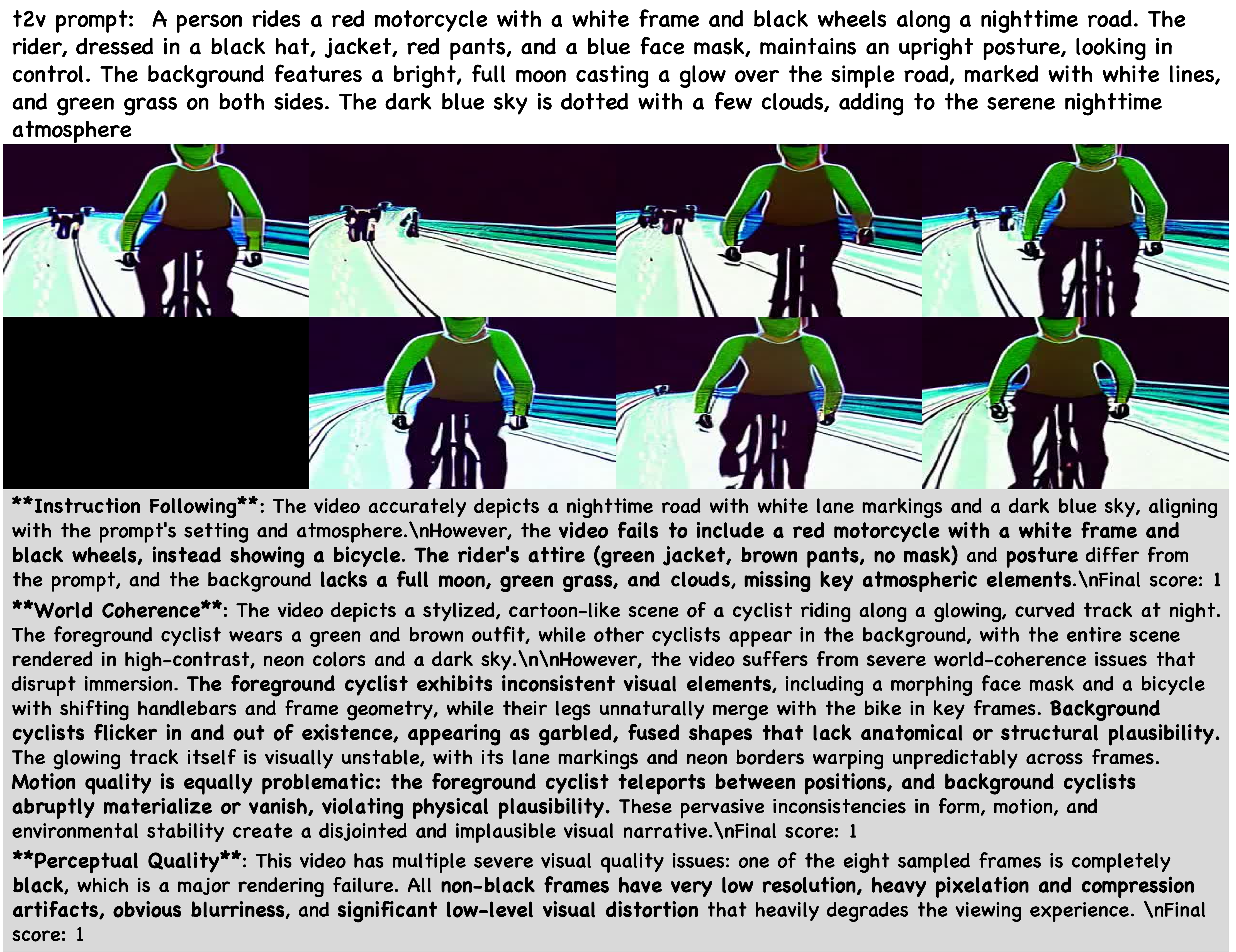}
    \caption{Qualitative example of FIRM-Video-8B evaluation (5)}
    \label{fig:case5}
\end{figure}

Figures~\ref{fig:case1}--\ref{fig:case5} present visualizations of FIRM-Video-8B's evaluations of samples with varying quality, serving as examples of quantitative analysis.

\clearpage

\clearpage

\clearpage

\clearpage

\end{document}